\documentclass{article}

\PassOptionsToPackage{numbers, compress}{natbib}
\newcommand{\bmX}{\mathcal{X}}
\newcommand{\bmY}{\mathcal{Y}}
\newcommand{\bmS}{\mathcal{S}}

\newcommand{\bmF}{\mathcal F}

\newcommand{\bmG}{\mathcal G}

\newcommand{\bmI}{\mathcal I}
\newcommand{\bmL}{\mathcal L}
\newcommand{\real}{\mathbb{R}}

\usepackage{microtype}
\usepackage{graphicx}
\usepackage{booktabs} 
\usepackage{caption}
\usepackage{subcaption}

\usepackage{amssymb, booktabs, multirow, amsmath}

\newtheorem{definition}{Definition}

\usepackage[final]{neurips_2021}

\usepackage[utf8]{inputenc} 
\usepackage[T1]{fontenc}    
\usepackage{hyperref}       
\usepackage{url}            
\usepackage{booktabs}       
\usepackage{amsfonts}       
\usepackage{nicefrac}       
\usepackage{microtype}      
\usepackage{xcolor}         
\usepackage{algorithmic}
\usepackage{algorithm}
\usepackage{cleveref}
\title{A Framework to Learn with Interpretation}

\author{%
  Jayneel Parekh \textsuperscript{\rm 1}, Pavlo Mozharovskyi \textsuperscript{\rm 1}, Florence d'Alch\'e-Buc \textsuperscript{\rm 1}\\
  \textsuperscript{\rm 1} LTCI, T\'el\'ecom Paris, \\ Institut Polytechnique de Paris, France \\
  \{jayneel.parekh,pavlo.mozharovskyi,florence.dalche\}@telecom-paris.fr
}

\begin{document}

\maketitle

\begin{abstract}

To tackle interpretability in deep learning, we present a novel framework to jointly learn a predictive model and its associated interpretation model. The interpreter provides both local and global interpretability about the predictive model in terms of human-understandable high level attribute functions, with minimal loss of accuracy. This is achieved by a dedicated architecture and well chosen regularization penalties. We seek for a small-size dictionary of high level attribute functions that take as inputs the outputs of selected hidden layers and whose outputs feed a linear classifier. We impose strong conciseness on the activation of attributes with an entropy-based criterion while enforcing fidelity to both inputs and outputs of the predictive model. A detailed pipeline to visualize the learnt features is also developed. Moreover, besides generating interpretable models \textit{by design}, our approach can be specialized to provide \textit{post-hoc} interpretations for a pre-trained neural network. We validate our approach against several state-of-the-art methods on multiple datasets and show its efficacy on both kinds of tasks.

\end{abstract}

\definecolor{purple}{RGB}{150,0,160}

\section{Introduction}




Interpretability in machine learning systems \cite{doshivelez2017,lipton2018,rudin2021} has recently attracted a large amount of attention. This is due to the increasing adoption of these tools in every area of automated decision-making, including critical domains such as law \cite{zeng2017}, healthcare \cite{stiglic2020} or defence. Besides robustness, fairness and safety, it is considered as an essential component to ensure trustworthiness in predictive models that exhibit a growing complexity. Explainability and interpretability are often used as synonyms in the literature, referring to the ability to provide human-understandable insights on the decision process. Throughout this paper, we opt for interpretability as in \cite{doshi2017towards} and leave the term explainability for the ability to provide logical explanations or causal reasoning, both requiring more sophisticated frameworks \cite{Dubois2014,Guidotti2019,Saralajew2019}.
To address the long-standing challenge of interpreting models such as deep neural networks \cite{Samekbook,Beaudouin2020,Barredo2020}, two main approaches have been developed in literature: \textit{post-hoc} approaches and ``\textit{by design} methods''.


Post-hoc approaches \cite{Baehrens2010,lime, shap, gradcam} generally analyze a pre-trained system locally and attempt to interpret its decisions. ``Interpretable by design'' \cite{cen,Adel2018} methods aim at integrating the interpretability objective into the learning process.  They generally modify the structure of predictor function itself or add to the loss function regularizing penalties to enforce interpretability. Both approaches offer different types of advantages and drawbacks. Post-hoc approaches guarantee not affecting the performance of the pre-trained system but are however criticized for computational costs, robustness and faithfulness of interpretations \cite{cost-saliency, kindermans2017unreliability, robust-saliency}. Interpretable systems by-design on the other hand, although preferred for interpretability, face the challenge of not losing out on performance. 

Here, we adopt another angle to learning interpretable models. As a starting point, we consider that prediction (computing $\hat{y}$ the model's output for a given input) and interpretation (giving a human-understandable description of properties of the input that lead to $\hat{y}$) are two distinct but strongly related tasks. On one hand, they do not involve the same criteria for the assessment of their quality and might not be implemented using the same hypothesis space. On the other hand, we wish that an interpretable model relies on the components of a predictive model to remain faithful to it. These remarks yield to a novel generic task in machine learning called Supervised Learning with Interpretation (SLI). SLI is the problem of jointly learning a pair of dedicated models, a predictive model and an interpreter model, to provide both interpretability and prediction accuracy. 
In this work, we present FLINT (Framework to Learn With INTerpretation) as a solution to SLI when the model to interpret is a deep neural network classifier. The interpreter in FLINT implements the idea that a prediction to be understandable by a human should be linearly decomposed in terms of attribute functions that encode high-level concepts as other approaches \cite{senn,ace}. However, it enjoys two original key features. First the high-level attribute functions leverage the outputs of chosen hidden layers of the neural network. Second, together with expansion coefficients they are jointly learnt with the neural network to enable local and global interpretations. By local interpretation, we mean a subset of attribute functions whose simultaneous activation leads to the model's prediction, while by global interpretation, we refer to the description of each class in terms of a subset of attribute functions whose activation leads to the class prediction. Learning the pair of models involves the minimization of dedicated losses and penalty terms. In particular, local and global interpretability are enforced by imposing a limited number of attribute functions as well as conciseness and diversity among the activation of these attributes for a given input. Additionally we show that FLINT can be specialized to post-hoc interpretability if a pre-trained deep neural network is available.

{\bf Key contributions:}
\vspace{-10pt}
 \begin{itemize}
     \setlength\itemsep{0pt}
     \item We present FLINT devoted to Supervised Learning with Interpretation with an original interpreter network architecture based on some hidden layers of the network. The role of the interpreter is to provide local and global interpretability that we express using a novel notion of relevance of concepts. 
     \item We propose a novel entropy and sparsity based criterion for promoting conciseness and diversity in the learnt attribute functions and develop a simple pipeline to visualize the encoded concepts based on previously proposed tools.  
     \item We present extensive experiments on 4 image classification datasets, MNIST, FashionMNIST, CIFAR10, QuickDraw, with a comparison with state-of-the-art approaches and a subjective evaluation study.
     \item Eventually, a specialization of FLINT to  post-hoc interpretability is presented while corresponding numerical results are deferred to supplements.
 \end{itemize}


\section{Related Works}

We emphasize here more on the methods relying upon a dictionary of high level attributes/concepts, a key feature of our framework. A synthetic view of this review is presented in the supplements to effectively view the connections and differences w.r.t wider literature regarding interpretability. 

%
{\bf Post-hoc interpretations.} 
Most works in literature focus on producing \textit{a posteriori} interpretations for pre-trained models via input attribution. They often consider the model as a black-box \cite{lime, shap, vibi, muse,Crabbe2020} or in the case of deep neural networks, work with gradients to generate saliency maps for a given input \cite{saliency, smoothgrad, gradcam, lrp}. Very few post-hoc approaches rely on high level concepts or other means of interpretations \cite{hendricks2016}. Methods utilizing high level concepts come under the subclass of concept activation vector (CAV)-based approaches. TCAV \cite{kim17tcav} proposed to utilize human-annotated examples to represent concepts in terms of activations of a pre-trained neural network. The sensitivity of prediction to these concepts is estimated to offer an explanation. ACE \cite{ace} attempts to automate the human-annotation process by super-pixel segmentation and clustering these segments based on their perceptual similarity where each cluster represents a concept. ConceptSHAP \cite{conceptshap} introduces the idea of ``completeness'' in ACE's framework. The CAV-based approaches already strongly differ from us in context of problem as they only consider post-hoc interpretations. TCAV generates candidate concepts using human supervision and not from the network itself. While ACE automates concept discovery, the concepts are less dynamic as by design they are associated to a single class and rely on being represented via spatially connected regions. Moreover, since ACE depends on using a CNN as perceptual similarity metric for image segments (regardless of aspect ratio, scale), it is limited in applicability (experimentally supported in supplement Sec. S.3).\\ 

{\bf Interpretable neural networks by design.} Most works from this class learn a single model by either modifying the architecture \cite{cen}, the loss functions \cite{icnn, chen2020}, or both \cite{xdnn, protodnn, senn, chen2019looks}. Methods like FRESH \cite{fresh} and INVASE \cite{invase} perform selection over raw input tokens/features. The selected input features then are used by the final prediction model. GAME \cite{game} shapes the learning problem as a co-operative game between predictor and interpreter. However, it learns a separate local interpreter for each sample rather than a single model. The above methods do not utilize high-level concepts for interpretation and offer local interpretations, with the exception of neural additive models \cite{nam}, which are currently only suitable for tabular data.\\
Self Explaining Neural Networks (SENN) \cite{senn} presented a generalized linear model wherein coefficients are also modelled as a function of input. The linear structure is to emphasize interpretability. SENN imposes a gradient-based penalty to learn coefficients stably and other constraints to learn human understandable features. 
Unlike SENN, to avoid trade-off between accuracy and interpretability in FLINT, we allow the predictor to be an unrestrained neural network and jointly learn the interpreter. Interpretations are generated at a local and global level using a novel notion of relevance of attributes. Moreover, FLINT can be specialized for generating post-hoc interpretations of pre-trained networks. \\
{\bf Known dictionary of concepts.} Some recent works have focused on different ways of utilizing a known dictionary of concepts for interpretability \cite{cme}, by transforming the latent space to align with the concepts \cite{chen2020} or by adding user intervention as an additional feature to improve interactivity \cite{bottleneck}. It should be noted that these methods are not comparable to FLINT or other interpretable networks by design as they assume availability of a ground truth dictionary of concepts for training.

\section{Learning a classifier and its interpreter with FLINT}
We introduce a novel generic task called \emph{Supervised Learning with Interpretation} (SLI). Denoting $\bmX$ the input space, and $\bmY$ the output space, we assume that the training set $\mathcal{S} = \{(x_i,y_i)_{i=1}^N \}$ is composed of $n$ independent realizations of a pair of random variables $(X,Y)$ defined over $\bmX \times \bmY$. SLI refers to the idea that the {\bf interpretation} task differs from the {\bf prediction} task and must be taken over by a dedicated model that depends on the predictive model to be interpreted. Let us call $\bmF$ the space of predictive models from $\bmX$ to $\bmY$. For a given model $f \in \bmF$, we denote $\bmG_f$ the family of models $g_f: \bmX \to \bmY$, that depend on $f$ and are devoted to its interpretation. For sake of simplicity, an interpreter $g_f \in \bmG_f$ is denoted $g$, omitting the dependency on $f$. With these assumptions, the empirical loss of supervised learning is revisited to include explicitly an interpretability objective besides the prediction loss yielding to the following definition.

{\bf Supervised Learning with Interpretation (SLI)}:\label{def:sli}
\begin{equation*}
\text{{\bf Problem 1}:}~ \arg \min_{f \in \bmF, g \in \bmG_{f}}  \bmL_{pred}(f,\bmS) +  \bmL_{int}(f,g,\bmS),
\end{equation*}
where $\bmL_{pred}(f,\bmS)$ denotes a loss term related to prediction error and $\bmL_{int}(f,g,\bmS)$ measures the ability of $g$ to provide interpretations of predictions by $f$. 

The goal of this paper is to address Supervised Learning with Interpretation when the hypothesis space $\bmF$ is instantiated to deep neural networks and the task at hand is multi-class classification. We present a novel and general framework, called Framework to Learn with INTerpretation (FLINT) that relies on (i) a specific architecture for the interpreter model which leverages some hidden layers of the neural network network to be interpreted, (ii) notions of local and global interpretation and (iii) corresponding penalties in the loss function. 
\subsection{Design of FLINT}
\begin{figure}[t]
\centering
\includegraphics[width=0.9\textwidth]{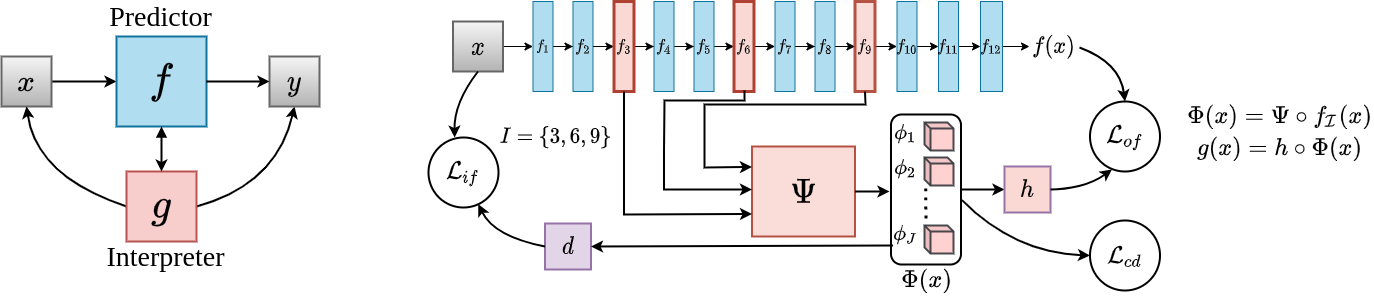}
\vspace{-3pt}
\caption{\textbf{(Left)} General view of FLINT. \textbf{(Right)} Instantiation of FLINT on a deep architecture.}
\label{fig_sys}
\end{figure}

All along the paper, we take $\bmX=\mathbb{R}^d$  and $\bmY=\{y \in \{ 0,1 \}^C, \sum_{j=1}^{C} y^j=1\}$, the set of  $C$ one-hot encoding vectors of dimension $C$.
We set $\bmF$ to the class of deep neural networks with $l$ hidden layers of respective dimension $d_1, \ldots, d_{l}$. Each element $f:\mathcal{X} \rightarrow \mathcal{Y}$ of $\bmF$ satisfies: $f = f_{l+1} \circ f_l \circ ... \circ f_1$ where $f_k: \mathbb{R}^{d_{k-1}} \rightarrow \mathbb{R}^{d_{k}}$, $d_0 = d, d_{l+1} = C, k = 1,...,l+1$ is the function implemented by layer $k$. 
A network $f$ in $\bmF$ is completely identified by its generic parameter $\theta_f$. 
As for the interpreter model $g \in \bmG_f$, we propose the following original architecture which exploits the outputs of chosen hidden layers of $f$. Denote $\mathcal{I} = \{i_1, i_2, ..., i_T\} \subset \{1,\ldots,l \}$ the set  of  indices specifying the intermediate layers of network $f$ to be accessed and chosen for the representation of input. We define $D= \sum_{t=1}^T d_{i_t}$. Typically these layers are selected from the latter layers of the network $f$. The concatenated vector of all intermediate outputs for an input sample $x$ is denoted as $f_{\bmI}(x) \in \mathbb{R}^D$.  
Given $f$ a network to be interpreted and a positive integer $J \in \mathbb{N}^*$, an {\bf interpreter network} $g$ computes the composition of a dictionary of attribute functions $\Phi: \bmX \to \real^J$ and an interpretable function $h: \real^J \to \bmY$.
\begin{equation}\label{eq:gen-g}
\forall x \in \bmX, g(x)= h \circ \Phi(x),
\end{equation}
In this work, we take: $h(\Phi(x)):=\text{softmax}(W^T\Phi(x))$ but other models like decision trees could be eligible.
The {\bf attribute dictionary} is composed of functions $\phi_j:\bmX \to \real^+,j=1, \ldots J$ whose non-negative images $\phi_j(x)$ can be interpreted as the activation of some high level attribute, i.e. a "concept" over $\bmX$. A key originality of the model lies in the fact that the attribute functions $\phi_j$ (referred to as attribute for simplicity) leverage the outputs of hidden layers of $f$ specified by $\mathcal{I}$: 
\begin{equation}
    \forall  j \in \{1, \ldots, J\}, \phi_j(x)= \psi_j \circ f_{\bmI}(x)
    \label{def:attribute}
\end{equation}
where each $\psi_j:\real^{D} \to \real^{+}$ operates on the accessed hidden layers. Here, the set of functions $\psi_j, j=1, \ldots J$ is defined to form a shallow network $\Psi$ (around 3 layers) whose output is $\Psi(f_{\bmI}(x))=\Phi(x)$ (example architecture in Fig. \ref{fig_sys}).
Interestingly, $\phi_j$ are defined over $\bmX$ and as a consequence can be interpreted in the input space which is the most meaningful for the user (see Sec.~\ref{sec:undertand}). For sake of simplicity, we denote $\Theta_g= (\theta_\Psi, \theta_h)$ the specific parameters of this model, while the parameters devoted to the computation of $f_{\bmI}(x)$ are shared with $f$.  
\subsection{Interpretation in FLINT}
The interpreter being defined, we need to specify its expected role and corresponding interpretability objective. In FLINT, interpretation is seen as an additional task besides prediction. We are interested by two kinds of interpretation, one at the global level that helps to understand which attribute functions are useful to predict a class and the other at the local level, that indicates which attribute functions are involved in prediction of a specific sample. As a preamble, note that, to interpret a local prediction $f(x)$, we require that the interpreter output $g(x)$ matches $f(x)$. When the two models disagree, we provide a way to analyze the conflictual data and possibly raise an issue about the confidence on the prediction $f(x)$ (see Supplementary Sec. S.2). 
To define local and global interpretation, we rely on the notion of relevance of an attribute.

Given an interpreter with parameter $\Theta_g= (\theta_\Psi, \theta_h)$ and some input $x$, the \textbf{relevance score} of an attribute $\phi_j$  is defined regarding the prediction $g(x)=f(x)=\hat{y}$.
Denoting $\hat{y} \in \bmY$ the index of the predicted class and $w_{j, \hat{y}} \in W$ the coefficient associated to this class, the contribution of attribute $\phi_j$ to unnormalized score of class $\hat{y}$ is $\alpha_{j, \hat{y}, x}=\phi_j(x).w_{j,\hat{y}}$. The relevance score is computed by normalizing contribution $\alpha$ as $r_{j, x} = \frac{\alpha_{j, \hat{y}, x}}{\max_i |\alpha_{i, \hat{y}, x}|}$. An attribute $\phi_j$ is considered as relevant for a local prediction if it is both activated and effectively used in the linear (logistic) model.
The notion of relevance of an attribute for a sample is extended to its "overall" importance in the prediction of any class $c$. This can be done by simply averaging relevance scores from local interpretations over a random subset or whole of the training set $\bmS$, where predicted class is $c$. Thus, we have: $r_{j, c} = \frac{1}{|\bmS_c|}\sum_{ x \in \mathcal{S}_{c}} r_{j, x}$, $\bmS_{c} = \{x \in \bmS | \hat{y} = c\}$.
Now, we can introduce the notions of local and global interpretations that the interpreter will provide.
\begin{definition}[Global and Local Interpretation]\label{def:intepretation}
For a prediction network $f$, the {\bf global interpretation}  $G(g,f)$ provided by an interpreter $g$, is the set of class-attribute pairs $(c, \phi_j)$ such that their global relevance $r_{j,c}$ is greater than some threshold $1/\tau, \tau > 1$.
A  {\bf local interpretation} for a sample $x$ provided by an interpreter $g$  of $f$ denoted $L(x,g,f)$ is the set of attribute functions $\phi_j$ with local relevance score $r_{j,x}$ greater than some threshold $1/\tau, \tau > 1$.
\end{definition}
It is important to note that these definitions do not prejudge the quality of local and global interpretations. Next, we convert desirable properties of the interpreter into specific loss functions.

\subsection{Learning by imposing interpretability properties}
\label{sec:loss}
Although converting desirable interpretability properties into losses is shared by several by-design approaches \cite{Carvalho2019,shap}, there is no consensus on these properties. We propose below a minimal set of penalties which are suitable for the proposed architecture and sufficient to provide relevant interpretations.

{\bf Fidelity to Output.} 
The output of the interpreter $g(x)$ should be "close" to $f(x)$ for any $x$. This can be imposed through a cross-entropy loss:
\begin{equation*}
    \mathcal{L}_{of}(f, g, \mathcal{S}) = - \sum_{x \in \mathcal{S}} h(\Psi(f_{\bmI}(x)))^T\log(f(x))
\end{equation*}

{\bf Conciseness and Diversity of Interpretations.}
For any given sample $x$, we wish to get a {\it small} number of attributes in its associated local interpretation. This property of \textit{conciseness} should make the interpretation easier to understand due to fewer attributes to be analyzed and promote the "high-level" character in the encoded concepts.  However, to encourage better use of available attributes we also expect activation of multiple attributes across many randomly selected samples. We refer to this property as {\it diversity}. This is also important to avoid the case of attribute functions being learnt as class exclusive (for eg. reshuffled version of class logits). To enforce these conditions we utilize notion of entropy defined for real vectors proposed by Jain et al \cite{subic} to solve problem of efficient image search. 
For a real-valued vector $v$, the entropy is defined as $\mathcal{E}(v) = -\sum_i p_i \log(p_i)$, $p_i = \exp(v_i)/(\sum_i \exp(v_i))$.

Conciseness is promoted by minimizing $\mathcal{E}(\Psi(f_{\bmI}(x)))$ and diversity is promoted by maximizing entropy of average $\Psi(f_{\bmI}(x))$ over a mini-batch. Note that this can be seen as encouraging the interpreter to find a sparse and diverse coding of $f_{\bmI}(x)$ using the function $\Psi$. Since entropy-based losses have inherent normalization, they do not constrain the magnitude of the attribute activation. This often leads to poor optimization. Thus, we also minimize the $\ell_1$ norm $\|\Psi(f_{\bmI}(x))\|_1$ (with hyperparameter $\eta$) to avoid it. Note that $\ell_1$-regularization is a common tool to encourage sparsity and thus conciseness, however we show in the experiments that entropy provides a more effective way.
\vspace{-3pt}
\begin{equation*}
    \mathcal{L}_{cd}(f, g, \mathcal{S}) = - \mathcal{E}(\bar{\Phi}_{\bmS}) + \sum_{x \in \mathcal{S}}\mathcal{E}(\Psi(f_{\bmI}(x))) + \sum_{x \in \mathcal{S}}\eta\|\Psi(f_{\bmI}(x))\|_1 \quad \text{with} \quad \bar{\Phi}_{\bmS} = \frac{1}{|\mathcal{S}|}\sum_{x \in \mathcal{S}} \Psi(f_{\bmI}(x))
\end{equation*}

{\bf Fidelity to Input.}
To encourage encoding high-level patterns related to input in $\Phi(x)$, we use a decoder network $d: \mathbb{R}^J \rightarrow \bmX$ that takes as input the dictionary of attributes $\Psi(f_{\bmI}(x))$ and reconstructs $x$. A similar penalty has previously been applied by \cite{senn}.
\begin{equation*}
     \mathcal{L}_{if}(f,g, d, \mathcal{S}) = \sum_{x \in \mathcal{S}}(d(\Psi(f_{\bmI}(x))) - x)^2
\end{equation*}
Note that one can modify $\bmL_{if}$ with other reconstruction losses as well (such as $\ell_1$-reconstruction).


Given the proposed loss terms, the loss for interpretability writes as follows:
\begin{equation*}
    \begin{split}
     \mathcal{L}_{int}(f, g , d, \mathcal{S})  = & \beta\mathcal{L}_{of}(f, g, \mathcal{S}) + \gamma\mathcal{L}_{if}(f, g, d, \mathcal{S})  + \delta\mathcal{L}_{cd}(f, g, \mathcal{S})
     \end{split}
\end{equation*}
where $\beta, \gamma, \delta$ are non-negative hyperparameters. The total loss to be minimized $\bmL = \bmL_{pred} + \bmL_{int}$, where the prediction loss, $\mathcal{L}_{pred}$, is the well-known cross-entropy loss.

Let us denote $\Theta=(\theta_f,\theta_d, \theta_\Psi, \theta_h)$ the parameters of these networks. Learning the models $f$, $\Psi$, $h$ and $d$ boils down to learning $\Theta$.
In practice, introducing all the losses at once often leads to very poor optimization. Thus, we follow the procedure described in Alg. \ref{alg:train}. We train the networks with $\mathcal{L}_{pred}, \mathcal{L}_{if}$ for the first two epochs and gain a reasonable level of accuracy. From the third epoch we introduce $\mathcal{L}_{of}$ and from the fourth epoch we introduce $\mathcal{L}_{cd}$ loss. 
\begin{algorithm}
\caption{Learning algorithm for FLINT}\label{alg:train}
\begin{algorithmic}[1]
\STATE \textbf{Input:} $\mathcal{S}$ \& parameters $\Theta=(\theta_f,\theta_d, \theta_\Psi, \theta_h)$ \& hyperparameters: $\beta_0, \gamma_0, \delta_0, \eta_0$ \& number of batches $B$ \& number of training epochs $N_{epoch}$.
\STATE Random initialization of parameter $\Theta_0$
\STATE $\Theta_1$ $\leftarrow$ Train ($\bmS, \Theta_0, \beta=0, \gamma_0, \delta=0, \eta=0, B, 2$) 
\COMMENT{\% Trains 2 epochs with $\bmL_{pred}, \bmL_{if}$}
\STATE $\Theta_2$ $\leftarrow$ Train ($\bmS, \Theta_1, \beta=\beta_0, \gamma_0, \delta=0, \eta=0, B, 1$)
\COMMENT{\% Trains 1 epoch with $\bmL_{pred}, \bmL_{if}, \bmL_{of}$}
\STATE $\hat{\Theta}$ $\leftarrow$ Train ($\bmS, \Theta_2, \beta_0, \gamma_0, \delta_0, \eta_0, B, N_{epoch}-3$)
\COMMENT{\% Trains with all losses}




\STATE \textbf{Output:} $\hat{\Theta}=(\hat{\theta}_f,\hat{\theta}_d, \hat{\theta}_\Psi, \hat{\theta}_h)$
\end{algorithmic}
\end{algorithm}

\section{Understanding encoded concepts in FLINT}\label{sec:undertand}


Once the predictor and interpreter are jointly learnt, interpretation can be given at the global and local levels as in Def. \ref{def:intepretation}. A key component to grasp the interpretations is to understand the concept encoded by each individual attribute function $\phi_j$, previously defined in Eq. \ref{def:attribute}. In this work, we focus on image classification and propose to represent an encoded concept as a set of visual patterns in the {\bf input space} which highly activate $\phi_j$. We present a pipeline to generate visualizations for global and local interpretation by adapting various previously proposed tools \cite{senn, vis_ijcv16}.
\begin{algorithm}
\caption{Visualization of global interpretation}\label{alg:global-v}
\begin{algorithmic}[1]
\STATE \textbf{Input:} (class,attribute):$(c,\phi_j)$ \& subset size:$l$ \& training set:$\bmS_n$ \&  AM+PI params:$(\lambda_{\phi},\lambda_{tv},\lambda_{bo})$
\STATE $\bmS_{c} = \{x|(x, c) \in \bmS_n\}$
\STATE $\textrm{MAS}(c,\phi_j,l) \leftarrow \arg \max_{\mathcal{M} \subset \bmS_c, | \mathcal{M} | = l} \sum_{x_i \in \mathcal{M}} \phi_j(x)$
\STATE FOR $x_k \in \textrm{MAS}(c,\phi_j,l)$
\STATE   $x_{vis}^k \leftarrow  \textrm{AM+PI}(x_k,\lambda_{\phi},\lambda_{tv},\lambda_{bo})$
\STATE ENDFOR
\STATE \textbf{Output:}$\{x_{vis}^1, \ldots, x_{vis}^l\}$, $\textrm{MAS}(c,\phi_j,l)$
\end{algorithmic}
\end{algorithm}




\paragraph{Visualization of global interpretation.} Given any class-attribute pair $(c, \phi_j)$ in the global interpretation $G(g,f)$, we first select a small subset of training samples from class $c$ that maximally activate $\phi_j$. This set of samples is referred to as maximum activating samples and denoted $\textrm{MAS}(c,\phi_j,l)$ where $l$ is the size of the subset (chosen as 3 in the experiments). 
Although, MAS reveal some information about the encoded concept, it might not be apparent what aspect of these samples causes activation of $\phi_j$. We thus propose further analyzing each element in MAS through tools that enhance the detected concept. This results in a much better understanding. The primary tool we employ is a modified version of activation maximization \cite{vis_ijcv16}, which we refer to as \textit{activation maximization with partial initialization} (AM+PI). 

Given a maximum activating sample $x' \in \textrm{MAS}(c,\phi_j,l)$, the key idea behind AM+PI is to synthesize appropriate input via optimization, that maximally activates $\phi_j$. We thus optimize a common activation maximization objective \cite{vis_ijcv16}:
$
    \arg\max_{x} \lambda_{\phi}\phi_j(x) - \lambda_{tv}\textrm{TV}(x) - \lambda_{bo}\textrm{Bo}(x)
$
, where $\textrm{TV}(.), \textrm{Bo}(.)$ are regularization terms. However, we initialize the procedure by low-intensity version of sample $x'$. This makes the optimization easier with the detected concept weakly present in the input. This also allows the optimization to ``fill'' the input to enhance the encoded concept. As an output, we obtain a map adapted to $x'$, that strongly activates $\phi_j$. Complete details of the AM+PI procedure are given in supplementary (Sec. S.2). Visualization of a class-attribute pair is summarized in Alg. \ref{alg:global-v}. Alternative useful tools are discussed in the supplementary (Sec. S.2).   

{\bf Local analysis.} Given any test sample $x_0$, one can determine its local interpretation $L(x_0, f, g)$, the set of relevant attribute functions accordingly to Def. \ref{def:intepretation}. To visualize a relevant attribute $\phi_j \in L(x_0, f, g)$, we can repeat the AM+PI procedure with initialization using low-intensity version of $x_0$ to enhance concept detected by $\phi_j$ in $x_0$. Note that the understanding built about any attribute function $\phi_j$ via global analysis, although not essential, can still be helpful to understand the generated AM+PI maps during local analysis, as these maps are generally similar.  

\section{Numerical Experiments for FLINT}

{\bf Datasets and Networks.} We consider $4$ datasets for experiments, MNIST \cite{MNIST}, FashionMNIST \cite{fmnist}, CIFAR-10 \cite{cifar10}, and a subset of QuickDraw dataset \cite{qdraw}. Additional results on CIFAR100 \cite{cifar10} (large number of classes) and Caltech-UCSD Birds-200-2011 \cite{cub} (large-scale images and large number of classes) are covered in supplementary (Sec. S.2.2). Our experiments include $2$ kinds of architectures for predictor $f$: (i) LeNet-based \cite{lenet} network for MNIST, FashionMNIST, and (ii) ResNet18-based \cite{resnet} network for QuickDraw, CIFAR. We select one intermediate layer for LeNet based network and two for ResNet based networks, from the last few convolutional layers as they are expected to capture higher-level features.  We set the number of attributes $J=25$ for MNIST, FashionMNIST, $J=24$ QuickDraw and $J=36$ for CIFAR. Further details about the QuickDraw subset, precise architecture, ablation studies about choice of hyperparameters (hidden layers, size of attribute dictionary, loss scheduling) and optimization details are available in supplementary (Sec. S.2).

\subsection{Quantitative evaluation of FLINT}
We evaluate and compare our model with other state-of-the-art systems regarding accuracy and interpretability.
The evaluation metrics for interpretability \cite{doshi2017towards} are defined to measure the effectiveness of the losses proposed in Sec. \ref{sec:loss}. Our primary method for comparison, wherever applicable, is SENN, as it is an interpretable network by design with same units for interpretation as FLINT. Other baselines include PrototypeDNN \cite{protodnn} for predictive performance, LIME \cite{lime} and VIBI \cite{vibi} for fidelity of interpretations. Implementation of our method is available on Github \footnote{\url{https://github.com/jayneelparekh/FLINT}}. Details for implementation of baselines are in supplementary (Sec. S.2).

{\bf Predictive performance of FLINT.}
\label{exp:acc}
\begin{table}
\scriptsize
\centering
\begin{tabular}{l c | c c c c c c c}
\toprule
\multicolumn{1}{c}{} &
\multicolumn{5}{c}{Accuracy (in \%)} & \multicolumn{3}{c}{Fidelity (in \%)} \\
\cmidrule[1pt](lr){2-6} 
\cmidrule[1pt](lr){7-9}
 & BASE-$f$ & SENN & PrototypeDNN & FLINT-$f$ & FLINT-$g$ & LIME & VIBI & FLINT-$g$\\
\midrule
MNIST & 98.9$\pm$0.1 & 98.4$\pm$0.1  & \textbf{99.2} & 98.9$\pm$0.2   & 98.3$\pm$0.2 & 95.6$\pm$0.4 & 96.6$\pm$0.7 & \textbf{98.7$\pm$0.1}\\
FashionMNIST & 90.4$\pm$0.1 & 84.2$\pm$0.3 & 90.0 & \textbf{90.5$\pm$0.2} & 86.8$\pm$0.4 & 67.3$\pm$1.3 & 88.4$\pm$0.3 & \textbf{91.5$\pm$0.1}\\
CIFAR10 & 84.7$\pm$0.3 & 77.8$\pm$0.7 & -- & \textbf{84.5$\pm$0.2} & 84.0$\pm$0.4 & 31.5$\pm$0.9 & 65.5$\pm$0.3 & \textbf{93.2$\pm$0.2}\\
QuickDraw & 85.3$\pm$0.2 & 85.5$\pm$0.4 & -- & \textbf{85.7$\pm$0.3} & 85.4$\pm$0.1 & 76.3$\pm$0.1 & 78.6$\pm$0.4 & \textbf{90.8$\pm$0.4}\\
\bottomrule
\vspace{1pt}
\end{tabular}
\caption[Results]{
Results for accuracy (in \%) and fidelity to FLINT-$f$ on different datasets. BASE-$f$ is system trained with just accuracy loss. FLINT-$f$, FLINT-$g$ denote the predictor and interpreter trained in our framework. Mean and standard deviation of 4 runs for each system are reported
}
\label{results_1}
\end{table}
There are two goals to validate related to predictor trained with FLINT (denoted FLINT-$f$), (i) Jointly training $f$ with $g$ and backpropagating loss term $\mathcal{L}_{int}$ does not negatively impact performance, and (ii) The achieved performance is comparable with other similar interpretable by-design models. For the former we compare the accuracy of FLINT-$f$ with same predictor architecture trained just with $\mathcal{L}_{pred}$ (denoted by BASE-$f$). For the latter goal we compare accuracy of FLINT-$f$ with accuracy of SENN and another interpretable network by design PrototypeDNN \cite{protodnn} that does not use input attribution for interpretations. Note that PrototypeDNN requires non-trivial changes to the model for running on more complex datasets, CIFAR10 and QuickDraw. To avoid any unfair comparison we skip these results. The accuracies are reported in Tab. \ref{results_1}. They indicate that training $f$ within FLINT does not result in any significant accuracy loss on any dataset. Also, FLINT is competitive with other interpretable by-design models.

{\bf Fidelity of Interpreter.}
The fraction of samples where prediction of a model and its interpreter agree, i.e predict the same class, is referred to as \textit{fidelity}. It is a commonly used metric to measure how well an interpreter approximates a model \cite{vibi, rope}. Note that, typically, for interpretable by design models, fidelity cannot be measured as they only consider a single model. However, to validate that the interpreter trained with FLINT (denoted as FLINT-$g$) achieves a reasonable level of agreement with FLINT-$f$, we benchmark its fidelity against a state-of-the-art black-box explainer VIBI \cite{vibi} and a traditional method LIME \cite{lime}. The results for this are provided in Tab. \ref{results_1} (last three columns). FLINT-$g$ consistently achieves higher fidelity. Even though it is not a fair comparison as other systems are black-box explainers and FLINT-$g$ accesses intermediate layers, they clearly show that FLINT-$g$ demonstrates high fidelity to FLINT-$f$.

{\bf Conciseness of interpretations.}
\begin{figure}
     \begin{subfigure}[b]{0.23\textwidth}
         \centering
         \includegraphics[width=\textwidth]{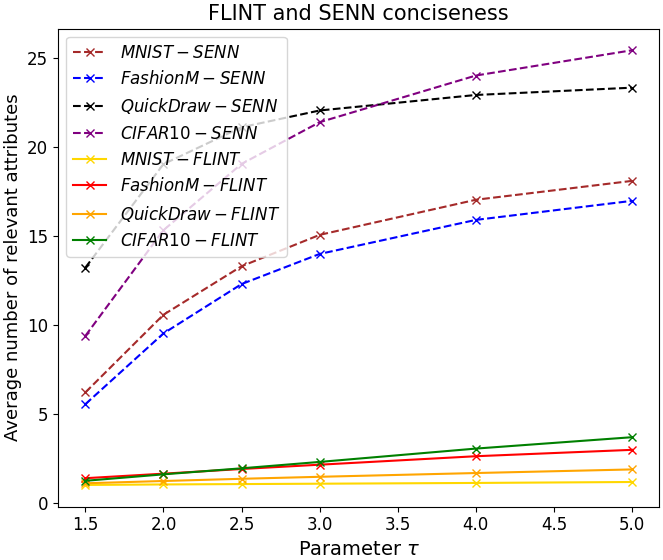}
         \caption{}
         \label{cns_1}
     \end{subfigure}
     ~
     \begin{subfigure}[b]{0.24\textwidth}
         \includegraphics[width=\textwidth]{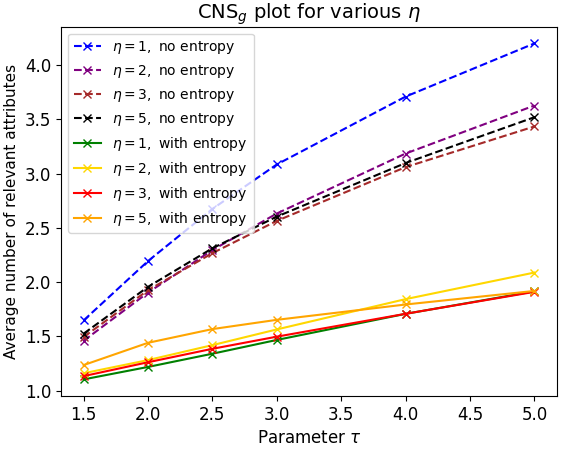}
         \caption{}
         \label{cns_2}
     \end{subfigure}
     ~
     \begin{subfigure}[b]{0.4\textwidth}
         \includegraphics[width=\textwidth]{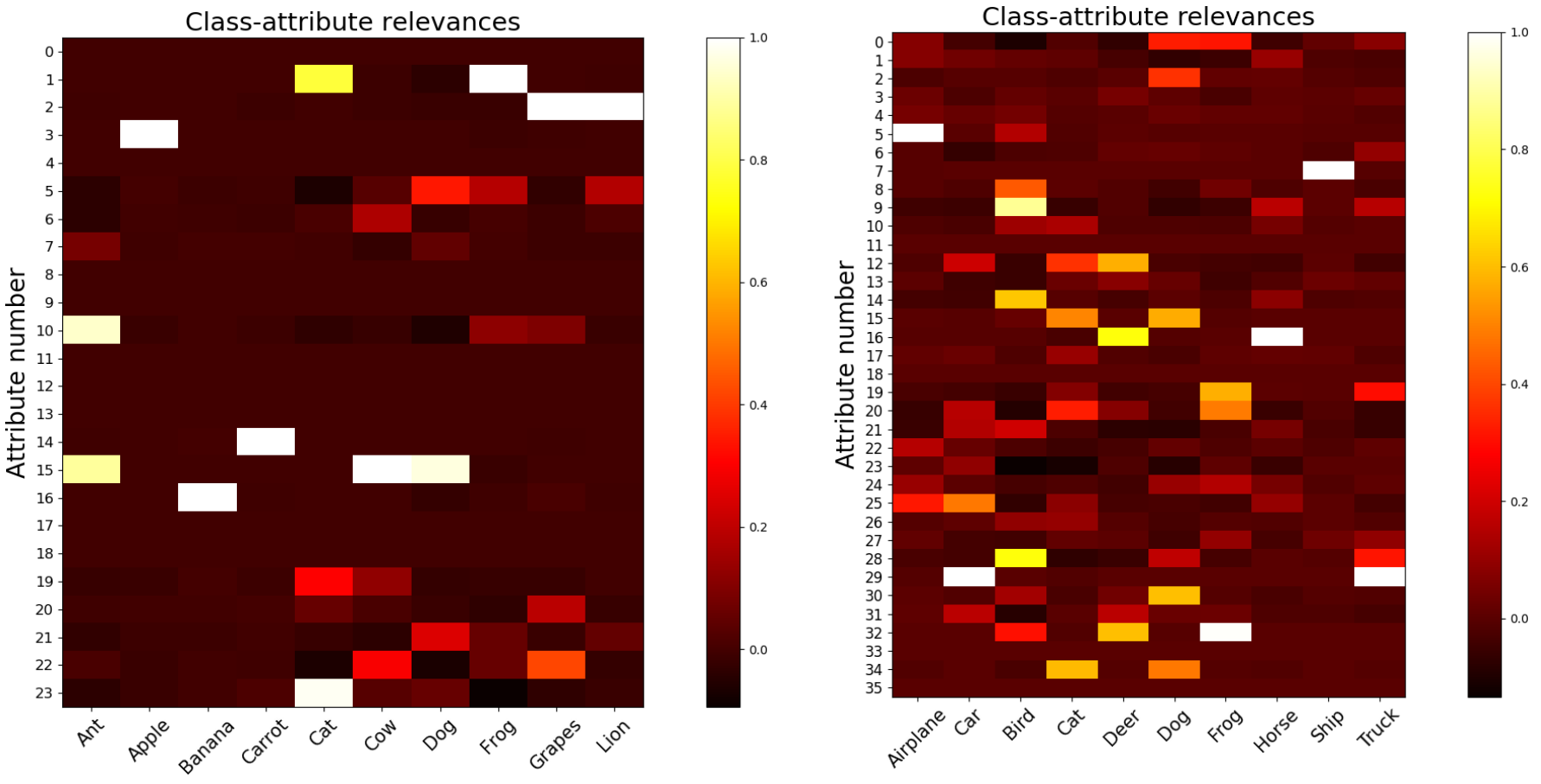}
         \caption{}
         \label{global_rel}
     \end{subfigure}
    \vspace{-3pt}
    \caption{(a) Conciseness comparison of FLINT and SENN. (b) Effect of entropy losses on conciseness of ResNet for QuickDraw for various $\ell_1$-regularization levels. (c) Global class-attribute relevances $r_{j,c}$ for QuickDraw (Left) and CIFAR10 (Right). 24 class-attribute pairs for QuickDraw and 32 pairs for CIFAR10 have relevance $r_{j,c} > 0.2$.}
    \label{fig:three graphs}
\end{figure}
We evaluate conciseness by measuring the average number of \textit{important} attributes in generated interpretations. For a given sample $x$, it can be computed as number of attributes $\phi_j$ with $r_{j,x}$ greater than a threshold $1/\tau, \tau > 1$, i.e. $\textrm{CNS}_{g,x} = |\{j: |r_{j,x}| > 1/\tau\}|$. For different thresholds $1/\tau$, we compute the mean of $\textrm{CNS}_{g,x}$ over test data to estimate conciseness of $g$, $\textrm{CNS}_g$. Lower conciseness indicates need to analyze a lower number of attributes on an average. SENN is the only other system for which this curve can be computed. We thus compare the conciseness of SENN with FLINT on all four datasets. Fig. \ref{cns_1} depicts the same. It can be easily observed that FLINT produces lot more concise interpretations compared to SENN. Moreover, SENN even ends up with majority of concepts being considered relevant for lower thresholds (higher $\tau$).

{\bf Entropy vs $\ell_1$ regularization.} We validate the effectiveness of entropy losses by computing conciseness curve at various levels of $\ell_1$ regularization strength, with and without entropy, for ResNet with QuickDraw. This is reported in Fig. \ref{cns_2}. The figure confirms that using the entropy-based loss is more effective in inducing conciseness of explanations compared to using just $\ell_1$-regularization, with the difference being close to use of  $1$ attribute less when entropy losses are employed.  

{\bf Importance of attributes.} Additional experiments evaluating meaningfulness of attributes by shuffling them and observing the effect (for FLINT and SENN) are given in supplementary (Sec. S.2).

\subsection{Qualitative analysis}


\begin{figure}[t!]
\centering
\includegraphics[width=0.99\textwidth]{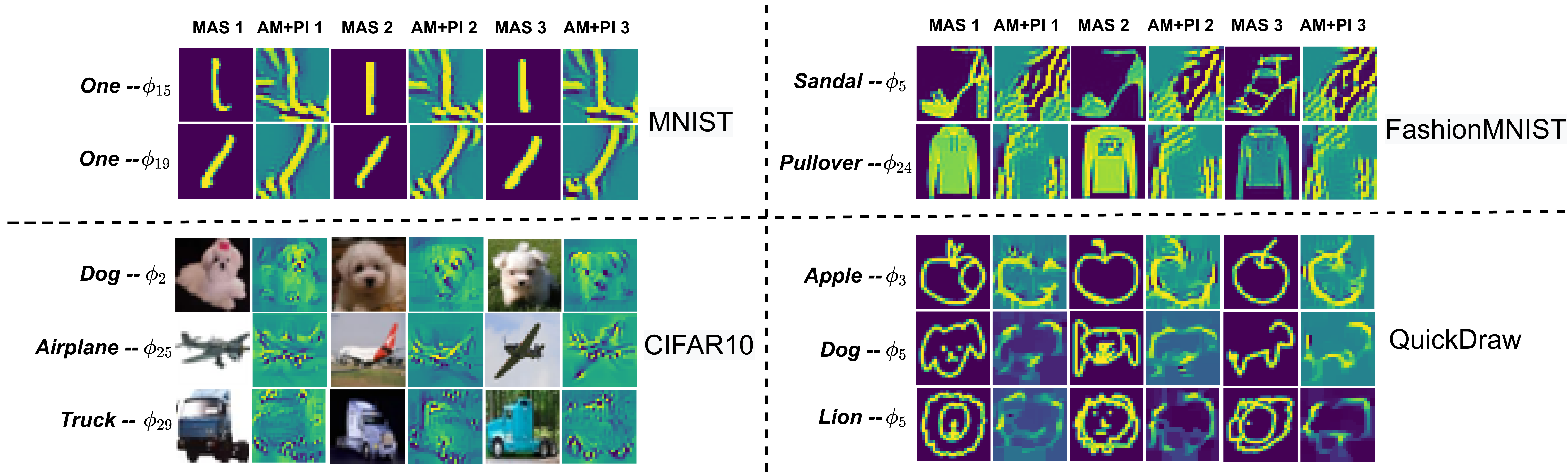}
\vspace{-4pt}
\caption{Example class-attribute pair analysis on all datasets, with global relevance $r_{j,c} > 0.2$. Each row contains 3 MAS with corresponding AM+PI outputs}
\label{global_viz}
\end{figure}

\vspace{-3pt}

\begin{figure}[t!]
 \centering
 \includegraphics[width=0.74\textwidth]{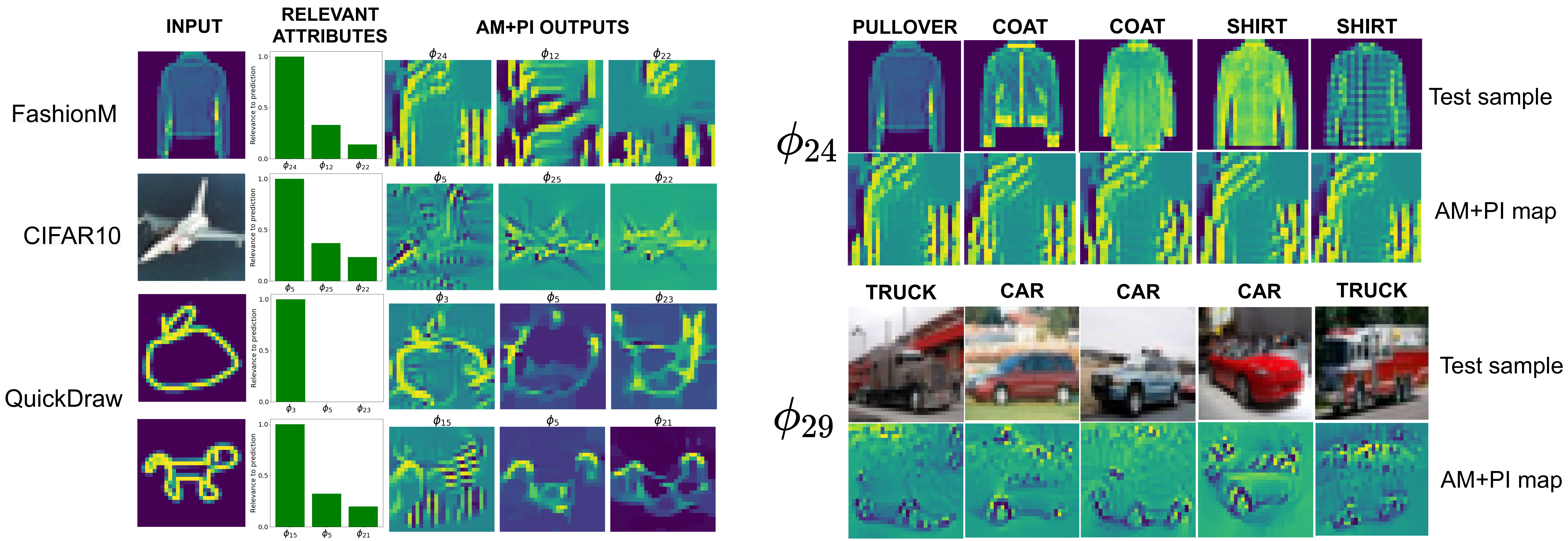}
 \vspace{-3pt}
 \caption{\textbf{(Left)} Local interpretations for test samples. Top 3 attributes with corresponding AM+PI output are shown. True labels for inputs are: Pullover, Airplane, Apple, Dog. \textbf{(Right)} Examples of attribute functions detecting same part across various test samples. For each sample, their relevance is greater than 0.8. True labels of samples indicated above them.}
 \label{local_int}
 \end{figure}


{\bf Global interpretation.} Fig. \ref{global_rel} depicts the generated global relevances $r_{j,c}$ for all class-attribute pairs on QuickDraw and CIFAR. Each class-attribute pair with `high' relevance needs to be analyzed as part of global analysis. Some example class-attribute pairs, with high relevance, are visualized in Fig.~\ref{global_viz}. For each pair we select MAS of size 3 and also show their AM+PI outputs. As mentioned before, simply analyzing MAS reveals useful information about the encoded concept. For instance, based on MAS, $\phi_{15}$, $\phi_{19}$ on MNIST, relevant for class `One', clearly seem to activate for vertical and diagonal strokes respectively. However, AM+PI outputs give deeper insights about the concept by revealing more clearly what parts of input activate an attribute function. For eg., while MAS indicate that $\phi_{5}$ on FashionMNIST activates for heels (one type of `Sandal'), $\phi_{2}$ on CIFAR10 activates for white dogs, it is not clear what part the attribute focuses on. AM+PI outputs indicate that $\phi_2$ focuses on the area around eyes and nose (the most enhanced regions), $\phi_{5}$ primarily detects a thin diagonal stroke of the heel surrounded by empty space. AM+PI outputs generally become even more important for attributes relevant for multiple classes. One such example is the function $\phi_5$ on QuickDraw, relevant for both `Dog' and `Lion'.  It activates for very similar set of strokes for all samples, as indicated by AM+PI maps. For `Dog' this corresponds to ears and mouth and for `Lion' it corresponds to the mane. Other such attribute functions in the figure include $\phi_{24}$ on FashionMNIST, relevant for `Pullover', `Coat' and `Shirt' which detects long sleeves and $\phi_{29}$ on CIFAR10, relevant for `Trucks', `Cars' and primarily detects wheels and parts of upper body. Further visualizations including those of other relevant classes for $\phi_{24}, \phi_{29}$ and global relevances are available in supplementary (Sec. S.2).

{\bf Local interpretation.} Fig. \ref{local_int} (left) displays the local interpretation visualizations for test samples. $f$ and $g$ both predict the true class in all the cases. We show the top $3$ relevant attributes to the prediction with their relevances and their corresponding AM+PI outputs. Based on the AM+PI outputs it can be observed that the attribute functions generally activate for patterns corresponding to the same concept as inferred during global analysis. This can be easily seen for attribute functions present in both Fig. \ref{global_viz}, \ref{local_int} (left). This is further illustrated by Fig. \ref{local_int} (right) where we illustrate AM+PI outputs for two attributes from Fig.  \ref{global_viz}. These functions are relevant for more than one class and detect the same concept across various test samples, namely long sleeves for $\phi_{24}$ and primarily wheels for $\phi_{29}$.  

\subsection{Subjective evaluation}
We conducted a \textit{survey based subjective evaluation} with QuickDraw dataset for FLINT with 20 respondents. We selected 10 attributes, covering 17 class-attribute pairs from the QuickDraw dataset. For each attribute we present the respondent with our visualizations (3 MAS and AM+PI outputs) for each of its relevant classes along with a textual description. We ask them if the description meaningfully associates to patterns in the AM+PI outputs. They indicate level of agreement with choices: Strongly Agree (SA), Agree (A), Disagree (D), Strongly Disagree (SD), Don't Know (DK). Descriptions were manually generated by our understanding of encoded concept for each attribute. 40\% incorrect descriptions were carefully included to ensure informed responses. These were forcefully related to the classes shown to make them harder to identify. \textbf{Results} -- for correct descriptions: 77.5\% -- SA/A, 10.0\% -- DK, 12.5\% -- D/SD. For incorrect descriptions: 83.7\% -- D/SD, 7.5\% -- DK, 8.8\% -- SA/A. These results clearly indicate that concepts encoded in FLINT's learnt attributes are understandable to humans. Survey details are given in supplementary (Sec. S.2).

\section{Specialization of FLINT to post-hoc interpretability}
\label{post-hoc}
While interpretability by design is the primary goal of FLINT, it can be specialized to provide a {\it post-hoc} interpretation when a classifier $\hat{f}$ is already available. The {\bf Post-hoc interpretation learning} (see for instance \citep{lime}) comes as a special case of SLI and is defined as follows. Given a classifier $\hat{f} \in \bmF$ and a training set $\bmS$, the goal is to build an interpreter of $\hat{f}$ by solving:
\vspace{-3pt}
\begin{equation*}
\text{{\bf Problem 2}:}~    \arg \min_{g \in \bmG_{\hat{f}}}   \bmL_{int}(\hat{f},g,\bmS).
\end{equation*}
With FLINT, we have $g(x)= h \circ \Phi(x)$ and $\Phi(x) = \Psi \circ \hat{f}_{\mathcal I}(x)$ for a given set of accessible hidden layers $\mathcal{I}$ and a attribute dictionary size $J$. Learning can be performed by specializing Alg. \ref{alg:train} with slight modification of replacing $\Theta$ as $\Theta = (\theta_{\Psi},\theta_h, \theta_d)$ while $\theta_{\hat{f}}$ is fixed and eliminating $\bmL_{pred}$ from training loss $\bmL$. 




{\bf Experimental results for post-hoc FLINT:} We validate this ability of our framework by interpreting fixed models trained only for accuracy, i.e, BASE-$f$ models from section \ref{exp:acc}. Even after not tuning the internal layers of $f$, the system is still able to generate high-fidelity and meaningful interpretations. Fidelity comparisons against VIBI, class-attribute pair visualizations and experimental details are available in supplementary (Sec. S.3).

\section{Discussion and Perspectives}

FLINT is a novel framework for learning a predictor network and its interpreter network with dedicated losses. It provides local and global interpretations in terms of high-level learnt attributes/concepts by relying on (some) hidden layers of the prediction network. This however leaves some under-explored questions about {\it faithfulness} of interpretations to the predictor. Defining faithfulness of an interpretation regarding a decision process \citep{senn} has not yet reached a consensus particularly in the case of post-hoc interpretability or when the two models, predictor and interpreter, differ \citep{yin21_faithfulness}.
Even though generating interpretations based on hidden layers of predictor ensures high level of faithfulness of the interpreter to the predictor, a complete faithfulness cannot be guaranteed since predictor and interpreter differ in their last part.
However if ensuring faithfulness by design is regarded as the primary objective, nothing stops the use of interpreter FLINT-$g$ as the final decision-making network. In this case, there is only one network and the so-called prediction network has only played the useful role of  providing relevant hidden layers. 

Retaining only the interpreter model additionally provides a compression of the predictor and can be relevant when frugality is at stake. Further works will investigate this direction and the enforcement of additional constraints on attribute functions to encourage invariance under various transformations. Eventually FLINT can be extended to other tasks or modalities other than images in particular by adapting the design of attributes and the pipeline to understand them.

\begin{ack}
This work has been funded by the research chair Data Science \& Artificial Intelligence for Digitalized Industry and Services of T\'{e}l\'{e}com Paris. This work also benefited from the support of French National Research Agency (ANR) under reference ANR-20-CE23-0028 (LIMPID project). The authors would like to thank Sanjeel Parekh for fruitful discussions and anonymous reviewers for their valuable comments and suggestions.
\end{ack}


\bibliography{references}
\bibliographystyle{plainnat}

\setcounter{section}{0}
\renewcommand{\thesection}{S.\arabic{section}}
\newpage
This supplementary is organized as follows:
\begin{itemize}
    \item Sec. S.1 contains a synthetic overview of various works in interpretability w.r.t FLINT.
    \item Sec. S.2 contains details and additional experiments regarding \textit{interpretability by-design} models.
    \item Sec. S.3 contains details and additional experiments regarding \textit{post-hoc} interpretations generated using FLINT.
    \item Sec. S.4 discusses the limitations of our proposed method.
    \item Sec. S.5 discusses the potential negative societal impact.
\end{itemize}

\section{Overview of related works}
To recap the properties of the methods exposed in Sec.~2 (main paper), we provide in Tab. \ref{comp} a synthetic view of the major properties of interpretable methods along three aspects.  \textit{Type} denotes if the method implements \textit{post-hoc} interpretations for a trained model or interpretable models \textit{by-design}). \textit{Scope} reflects the ability of the approach to provide interpretation of decisions for individual samples (\textit{Local}) or to understand the model as a whole (\textit{Global}). \textit{Means} denotes the units in which the interpretations are generated. Categories include raw input features, a simplified representation of input, logical rules, prototypes, high-level concepts. 
\begin{table}[h!]
\centering
\begin{tabular}{l c c c} 
\toprule
System & Means & Type & Scope\\
\toprule
LIME, SHAP & Simplified input & Post-hoc & Local+Global\\
Gradient based & Raw input & Post-hoc & Local\\
VIBI, L2X & Raw input & Post-hoc & Local\\
Anchors & Logical rules & Post-hoc & Local\\
\midrule
ICNN & Raw input & By-design & Local\\
INVASE & Raw input & By-design & Local\\
CEN, GAME & Simplified input & By-design & Local\\
PrototypeDNN & Prototypes & By-design & Local\\
\midrule
CAV-based & Concepts (External) & Post-hoc & Local+Global\\
SENN & Concepts (Learnt) & By-design & Local\\
FLINT & Concepts (Learnt) & Both & Local+Global\\

\bottomrule
\end{tabular}
\caption[Comparison]{Various interpretability systems and their properties.}
\label{comp}
\end{table}

\begin{itemize}
    \item LIME, SHAP:  Local Interpretable Model-agnostic Explanations \cite{lime}, SHapley Additive exPlanations \cite{shap}.
    
    \item VIBI, L2X: Variational Information Bottleneck for Interpretation \cite{vibi}, Learning to Explain \cite{l2x}.
    
    \item ICNN: Interpretable CNN \cite{icnn}.
    
    \item INVASE: Instance-Wise Variable Selection using Neural Networks \cite{invase}.
    
    \item CEN, GAME: Contextual Explanation Networks \cite{cen}, Game-theoretic transparency\cite{game}.
    
    \item PrototypeDNN: \cite{protodnn}.
    
    \item Anchors: \cite{anchors}.
    
    \item CAV-based: Testing with Concept Activation Vectors (TCAV) \cite{kim17tcav}, Towards Automatic Concept-based Explanations (ACE) \cite{ace}, ConceptSHAP \cite{conceptshap}.
    
    \item SENN: Self Explaining Neural Networks \cite{senn}.
\end{itemize}

\section{Interpretability by design: Additional information and experiments}

We cover all the implementation details in Sec.~\ref{exp_details}, including network architectures, choice of hyperparameters, optimization procedures, resource consumption. Experiments on CIFAR100 and CUB-200 are detailed in Sec.~\ref{cub_exp}. Additional analysis including ablation studies and visualizations for attributes are available in Sec.~\ref{analysis_viz}. We also present other useful tools for analysis in Sec.~\ref{other_tools}. Baseline implementations are discussed in Sec.~\ref{baselines}. Details about the subjective evaluation, including the form link are available in Sec. ~\ref{sub_eval}. Note that the experiments with ACE are deferred to Sec. \ref{ace_exp}.


\subsection{Implementation details}
\label{exp_details}

\subsubsection{Network architectures}
\label{net_archs}

\begin{figure*}[t!]
\centering
\includegraphics[width=0.9\textwidth]{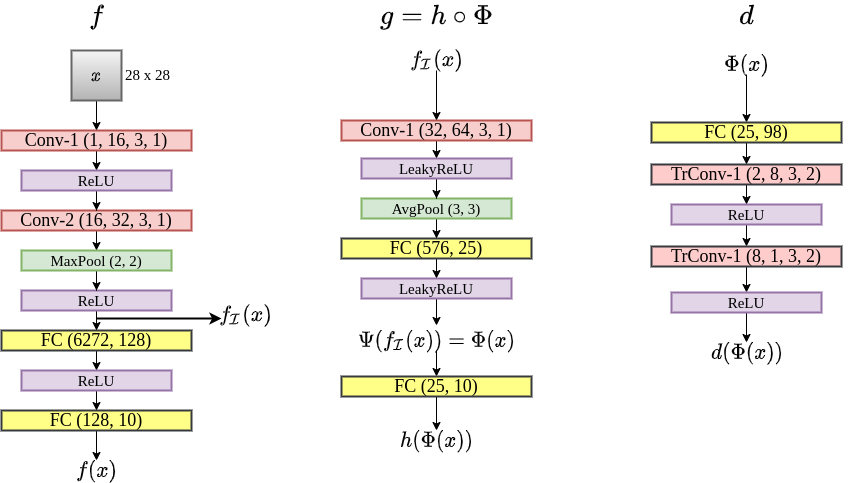} 
\caption{Architecture of networks based on LeNet \cite{lenet}. Conv (a, b, c, d) and TrConv (a, b, c, d) denote a convolutional, transposed convolutional layer respectively with number of input maps a, number of output maps b, kernel size c $\times$ c and stride size d. FC(a, b) denotes a fully-connected layer with number of input neurons a and output neurons b. MaxPool(a, a) denotes window size a $\times$ a for the max operation. AvgPool(a, a) denotes the output shape a $\times$ a for each input map }
\label{arch_1}
\end{figure*}

\begin{figure*}[t!]
\centering
\includegraphics[width=0.98\textwidth]{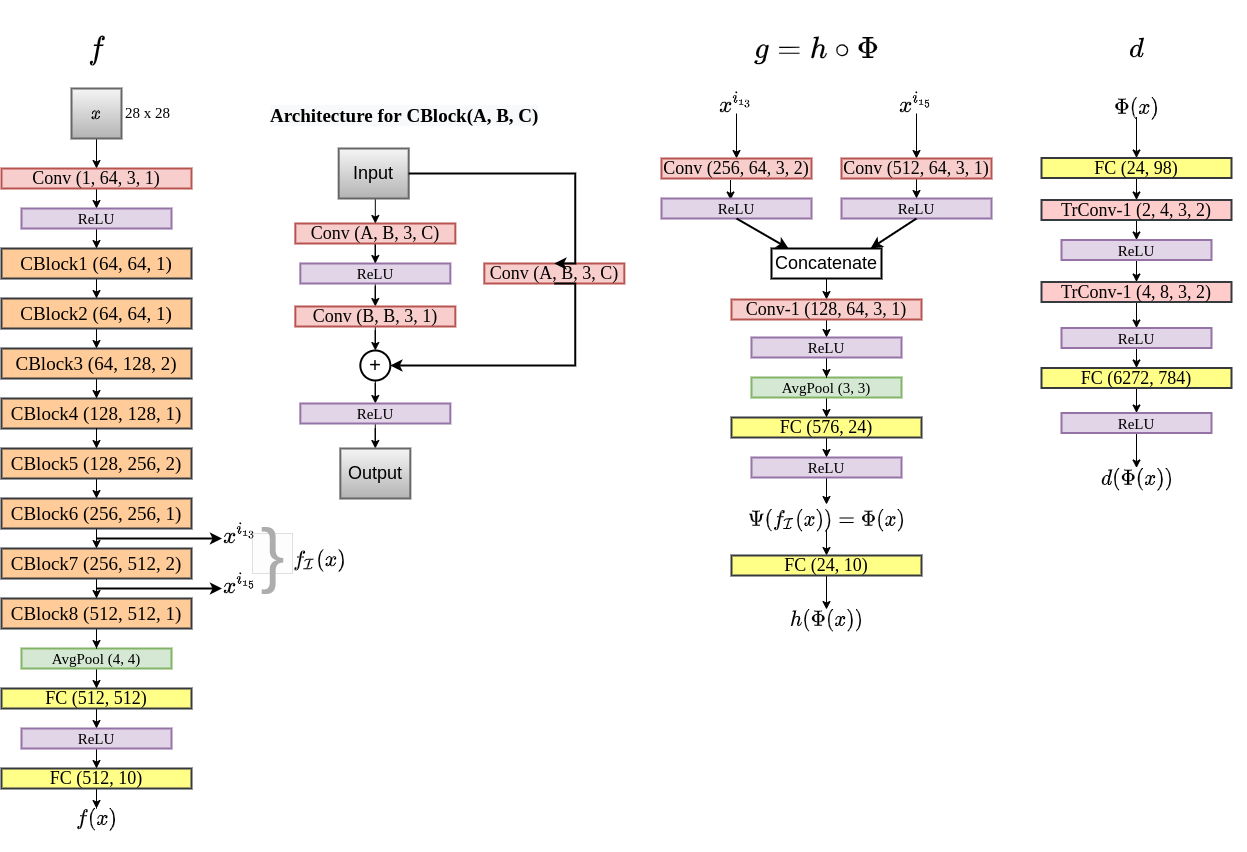} 
\caption{Architecture of networks for experiments on QuickDraw with network based on ResNet \cite{resnet}. Conv (a, b, c, d) and TrConv (a, b, c, d) denote a convolutional, transposed convolutional layer respectively with number of input maps a, number of output maps b, kernel size c $\times$ c and stride size d. FC(a, b) denotes a fully-connected layer with number of input neurons a and output neurons b. AvgPool(a, a) denotes the output shape a $\times$ a for each input map. Notation for CBlock is explained in the figure.}
\label{arch_2}
\end{figure*}

\paragraph{Predictor} Fig.~\ref{arch_1} and \ref{arch_2} depict the architectures used for experiments with predictor architecture based on LeNet \cite{lenet} (on MNIST, Fashion-MNIST) and ResNet18 (on CIFAR10, QuickDraw) \cite{resnet} respectively.

\paragraph{Interpreter} 
The architecture of interpreter $g = h \circ \Phi$ and decoder $d$ for MNIST, FashionMNIST are shown in Fig.~\ref{arch_1}. Corresponding architectures for QuickDraw are in Fig.~\ref{arch_2}. For CIFAR-10, the interpreter architecture is almost exactly the same as QuickDraw, with only difference being output layer for $\Phi(x)$, which contains 36 attributes instead of 24. The decoder $d$ also contains corresponding changes to input and output FC layers, with 36 dimensional input in first FC layer and $3072$ dimensional output in last FC layer.

The choice of selection of intermediate layers is an interesting part of designing the interpreter. In case of LeNet, we select the output of final convolutional layer. For ResNet, while we tend to select the intermediate layers from the latter convolutional layers, we do not select the last convolutional block (CBlock 8) output. This is mainly because empirically, when selecting the output of CBlock 8, the attributes were trivially learnt, with only one attribute activating for any sample and attributes exclusively activating for a single class. The hyperparameters are much harder to tune to avoid this scenario. Thus we selected two outputs from CBlock 6, CBlock 7 as intermediate layers. The layers in the interpreter itself were chosen fairly straightforwardly with 1-2 conv layers followed by a pooling and fully-connected layer.

\subsubsection{QuickDraw subset and pre-processing}
\paragraph{QuickDraw.} We created a subset of QuickDraw from the original dataset \cite{qdraw}, by selecting 10000 random images from each of 10 classes: 'Ant', 'Apple', 'Banana', 'Carrot', 'Cat', 'Cow', 'Dog', 'Frog', 'Grapes', 'Lion'. We randomly divide each class into $8000$ training and $2000$ test images.

\paragraph{Input pre-processing.} For MNIST, FashionMNIST and QuickDraw, we use the default images with pixel values in range $[0, 1]$. No data augmentation is performed. For CIFAR-10 we apply the most common mean and standard deviation normalization. The training data is generated by randomly cropping a $32 \times 32 \times 3$ image after padding the original images by zeros (size of padding is 2).

\subsubsection{Hyperparameter settings} 
\label{hyperparams_text}

\begin{table*}[ht]
\small
\centering
\begin{tabular}{l c c c c} 
\toprule
Variable & MNIST & FashionM & CIFAR10 & QuickDraw\\
\toprule
$N_{epoch}$ -- Number of training epochs & 12 & 12 & 25 & 12\\


$\beta$ -- Weight for $\bmL_{of}$ & 0.5 & 0.5 & 0.6 & 0.1\\
$\gamma$ -- Weight for $\bmL_{if}$ & 0.8 & 0.8 & 2.0 & 5.0\\
$\delta$ -- Weight for $\bmL_{cd}$ & 0.2 & 0.2 & 0.2 & 0.1\\
$\eta$ -- Relative strength of $\ell_1$-regularization  & 0.5 & 0.5 & 1.0 & 3.0\\

\bottomrule
\end{tabular}
\caption[Results]{Hyperparameters for FLINT
}
\label{hyperparams}
\end{table*}

\begin{table}
\centering
\begin{tabular}{l c c c c} 
\toprule
 & $\eta=1$ & $\eta=2$ & $\eta=3$ & $\eta=5$\\
\toprule
no entropy & 92.7 & 90.4 & 91.2 & 84.2\\
\midrule
with entropy & 91.2 & 90.7 & 90.8 & 82.9\\
\bottomrule
\end{tabular}
\caption[Results]{Fidelity (in \%) variation for $\eta$ and entropy losses for QuickDraw. $\delta=0.1$ is fixed 
}
\label{results_3}
\end{table}

Tab. \ref{hyperparams} reports the setting of our hyperparameters for different datasets. We briefly discuss here our method to tune the different weights.   

We varied $\gamma$ between 0.8 to 20 for all datasets, and stopped at a value for which the $\bmL_{if}$ loss seemed to optimize well (value dropped by at least 50\% compared to the start). For MNIST and FashionMNIST, the first value, 0.8 worked well. For the others, $\gamma$ needed to be increased so that the autoencoder worked well. Too high $\gamma$ might result in failed optimization due to exploding gradients. 

The variation of $\beta$ was based on two indicators: (i) The system achieves high fidelity, for eg. at least 90\%, so too small $\beta$ can't be chosen, (ii) For high $\beta$, the attributes become class-exclusive with only one attribute activating for a sample and result in high $\bmL_{if}$. Thus, $\beta$ was varied to get high fidelity and avoiding second scenario. $\beta=0.5$ worked well for MNIST, FashionMNIST. For QuickDraw, we needed to decrease $\beta$ because of second scenario. 

The system is fairly robust to choice of $\delta$, $\eta$. Too high $\ell_1$ regularization results in loss of fidelity (Tab. \ref{results_3}). These values were mostly heuristically chosen, and small changes to them do not cause much difference to training. We kept the effect of entropy low for ResNet because of its very deep architecture and high computational capacity of intermediate layers which can easily sway attributes to be class-exclusive. 

\subsubsection{AM+PI procedure}

In our case this optimization problem for an attribute $j$ is:
    \begin{equation*}
        \arg\max_{x} \lambda_{\phi}\phi_j(x) - \lambda_{tv}\textrm{TV}(x) - \lambda_{bo}\textrm{Bo}(x)
    \end{equation*}
    where $\textrm{TV}(x)$ denotes total variation of $x$ and $\textrm{Bo}(x)$ promotes boundedness of $x$ in a range. We fix parameters for AM+PI for MNIST, FashionMNIST, QuickDraw as $\lambda_{\phi}=2, \lambda_{tv}=6, \lambda_{bo}=10$ and $\lambda_{\phi}=2, \lambda_{tv}=20, \lambda_{bo}=20$ for CIFAR10. For each sample $x_0$ to be analyzed, we analyze input for this optimization as $0.3x_0$ for MNIST, FashionMNIST, QuickDraw and as $0.4x_0$ for CIFAR10. For optimization, we use Adam with learning rate $0.05$ for $300$ iterations, halving learning rate every $50$ iterations.

\subsubsection{Optimization and Runs} The models are trained for $12$ epochs on MNIST, FashionMNIST and QuickDrawm and for 25 epochs on CIFAR-10. We use Adam \cite{adam} as the optimizer with fixed learning rate $0.0001$ and train on a single NVIDIA-Tesla P100 GPU. Implementations are done using PyTorch \cite{pytorch}.

\textbf{Number of runs}: For the accuracy and fidelity results in the main paper, we have reported mean and standard deviation for 4 runs with different seeds for each system. The conciseness results are computed by averaging conciseness of 3 models for each reported system.


\subsubsection{Resource consumption}

Compared to $f$, $\Psi, h$ and $d$ have fewer parameters. For networks shown in Fig.~\ref{arch_1}, the LeNet based predictor has around 800,000 trainable parameters, interpreter $g$ contains 70,000 parameters, decoder $d$ contains 3000 parameters. For networks in Fig.~\ref{arch_2}, ResNet based predictor contains 11 million parameters, interpreter $g$ contains 530,000 parameters, and decoder $d$ contains 4.9 million parameters (almost all of them in the last FC layer). In terms of space, FLINT occupies more storage space according to the decoder, but is still of comparable size to that of only storing predictor. 

\paragraph{Training time}

In terms of training time consumption there is lesser difference  when $f$ is a very deep network, due to all networks $\Psi, h, d$ being much shallower (lesser number of layers) than $f$. For eg. on both CIFAR-10, QuickDraw, FLINT consumes just around 10\% more time for training compared to training just the predictor (BASE-$f$). The difference is more pronounced on with shallower $f$ where $\Psi, h, d$ also have comparable number of layers to $f$. Training BASE-$f$ on MNIST consumes 50\% less time compared to FLINT.

We compare the average training times (for four runs) for SENN and FLINT in Tab. \ref{time}. Each model is trained for the same number of epochs, on the same computing machine (1 NVIDIA Tesla P100 GPU). It is clear that SENN requires significantly more time to train. This is primarily because of gradient of output w.r.t input being part of their loss function. Thus the computational graph for a forward pass is twice as big as their model architecture and followed by a backward pass through the bigger graph. 

\begin{table}[ht]
\small
\centering
\begin{tabular}{l c c c} 
\toprule
Dataset & SENN & FLINT\\
\toprule
MNIST & 2311 & \textbf{518}\\
FashionMNIST & 2333 & \textbf{519} \\ 
CIFAR-10 & 10210 & \textbf{1548}\\
QuickDraw & 10548 & \textbf{1207}\\
\bottomrule
\end{tabular}
\caption[Results]{Training times for FLINT and SENN (in seconds)
}
\label{time}
\end{table}

\vspace{-7pt}

\subsection{Experiments on CIFAR-100 and CUB-200}
\label{cub_exp}

We also demonstrate the ability of the system to handle more complex datasets by experimenting with CIFAR100 \cite{cifar10} and Caltech-UCSD-200 (CUB-200) fine-grained Bird Classification dataset \cite{cub}. CIFAR100 contains 100 classes with 500 training and 100 testing samples per class (image size $32 \times 32 \times 3$). CUB-200 contains 11,788 images of 200 categories of birds, 5,994 for training and 5,794 for testing. We scale each sample in CUB-200 to size $224 \times 224 \times 3$. We also don't crop using the bounding boxes and use the full images for training and testing.  

Compared to our earlier experiments, we make two key changes to the framework, (i) Increase size of dictionary of attribute functions to accommodate larger images/number of classes, (ii) Modify architecture of decoder $d$ with more upsampling and convolutional layers. For CIFAR100, the same architectures for $f$ and $g$ as on CIFAR10 is used, but with $J=72$. We apply random horizontal flip as additional augmentation and train for $51$ epochs. For CUB-200, we use the ResNet18 \cite{resnet} for large-scale images as predictor architecture. We use $J=180$, and apply random horizontal flip and random cropping of zero-padded image as data augmentation. The predictor is initialized with network pretrained on ImageNet and trained for 50 epochs. For both datasets, we do not vary the other hyperparameters much compared to experiments on CIFAR10. The hidden layers accessed are same for both. The hyperparameters of the interpretability loss remain unchanged for CIFAR100 and for CUB-200 we increase $\beta$ and $\gamma$ to $1.0$ and $3.0$, respectively.

\begin{figure}[t]
\centering
\includegraphics[width=0.7\textwidth]{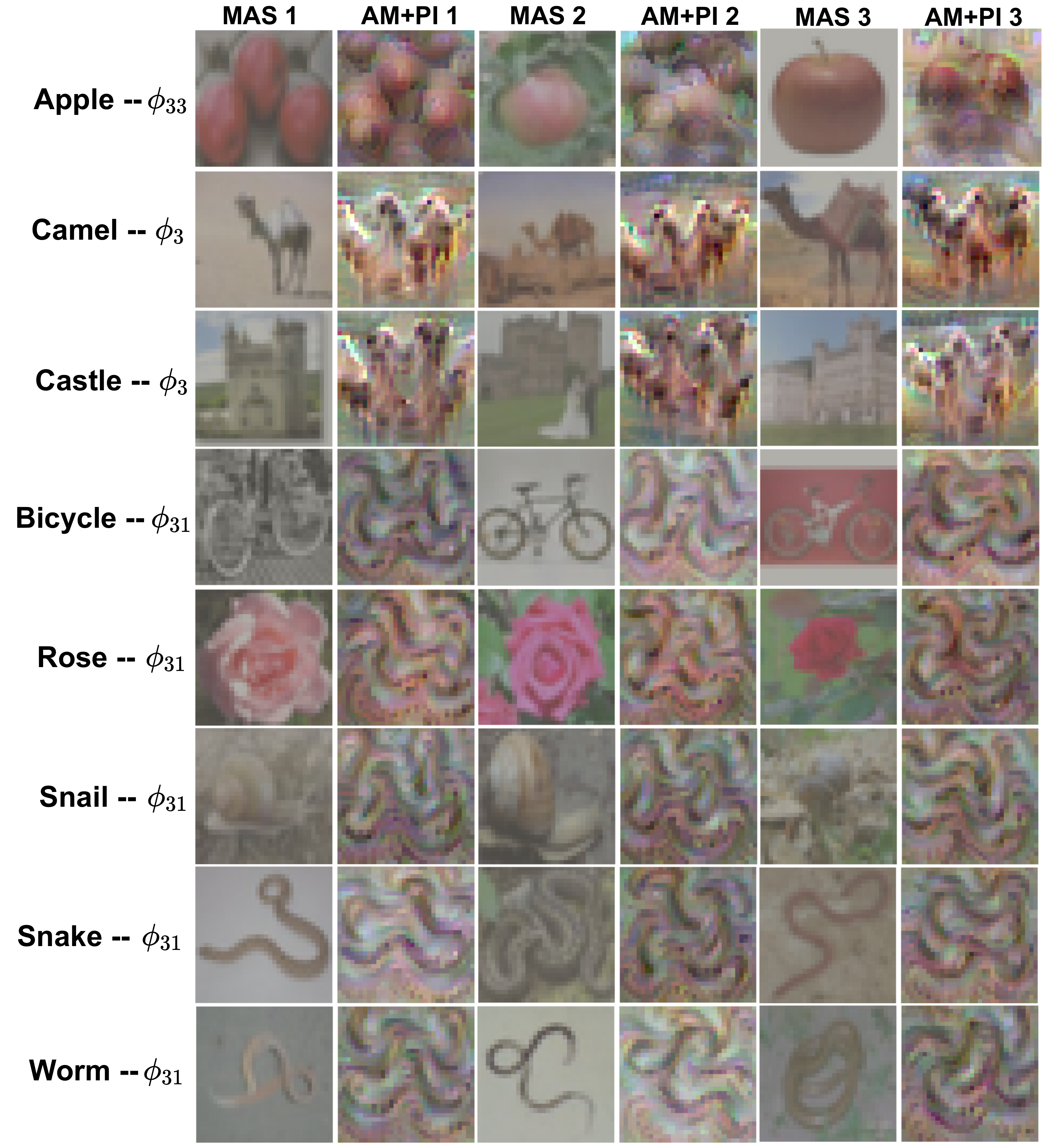} 
\caption{Sample class-attribute visualizations for CIFAR100. Three MAS and their corresponding AM+PI outputs are shown.}
\label{cifar100_global_viz}
\end{figure}

\begin{figure}[!htb]
    \centering
    \includegraphics[width=1.11\textwidth]{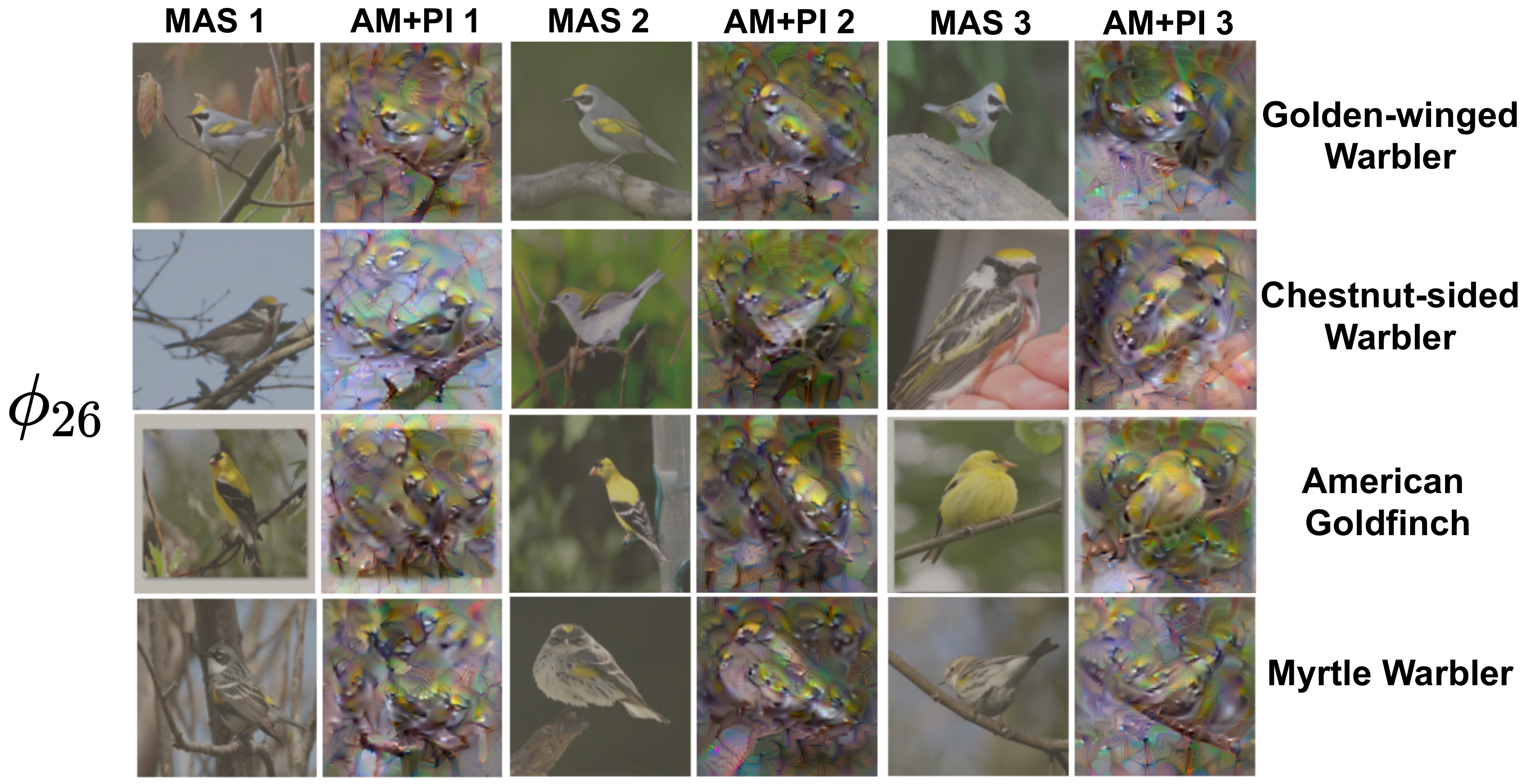}
    \caption{Relevant class-attribute pairs on CUB-200 with attribute $\phi_{26}$. Each row gives visualization for a relevant class of the attribute with three MAS and corresponding AM+PI outputs.}
    \label{cub200_global_1}
\end{figure}

\begin{figure}[!htb]
    \centering
    \includegraphics[width=1.11\textwidth]{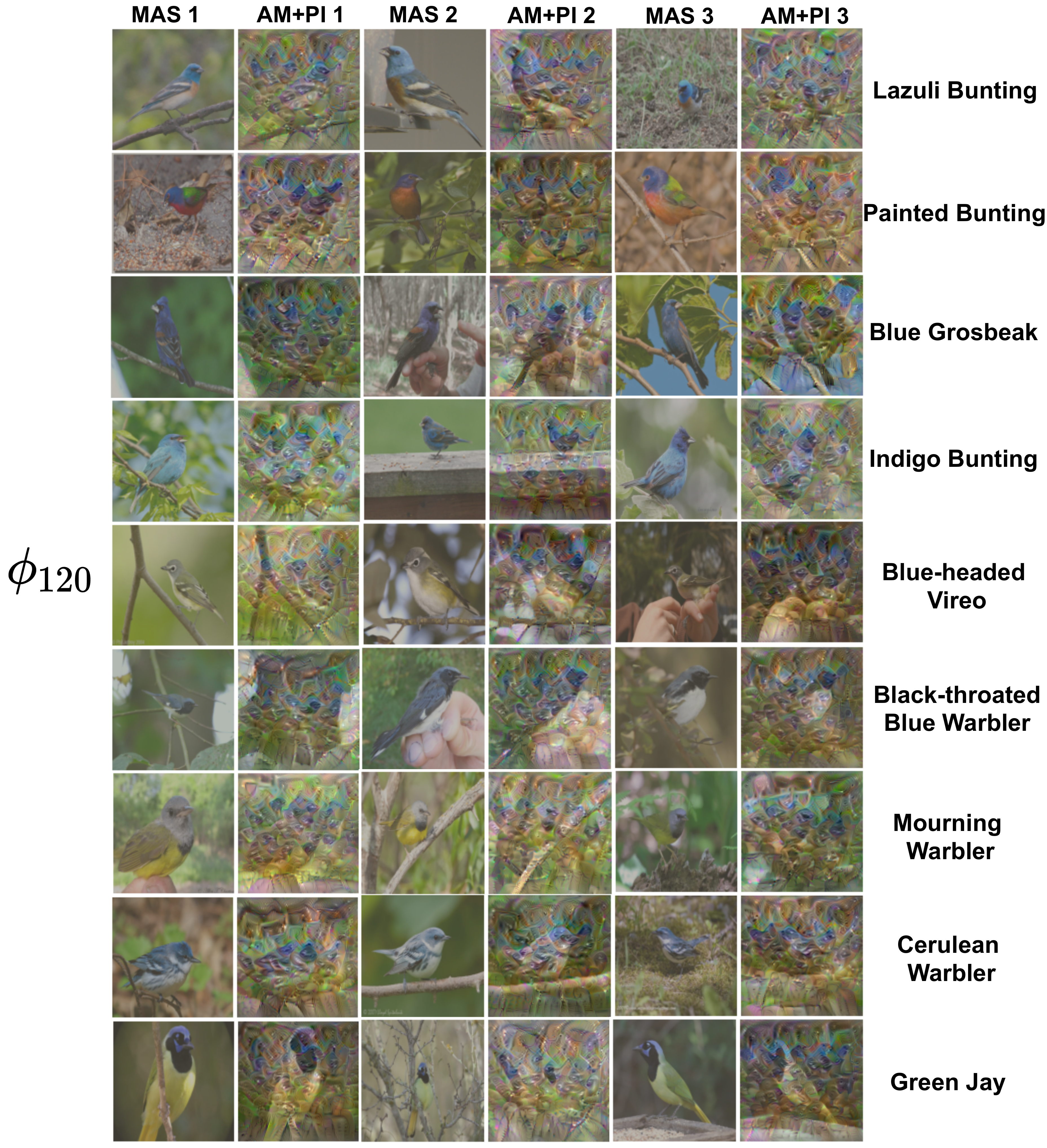}
    \caption{Relevant class-attribute pairs on CUB-200 with attribute $\phi_{120}$. Each row gives visualization for a relevant class of the attribute with three MAS and corresponding AM+PI outputs.}
    \label{cub200_global_3}
\end{figure}

\begin{figure}[!htb]
    \centering
    \includegraphics[width=1.11\textwidth]{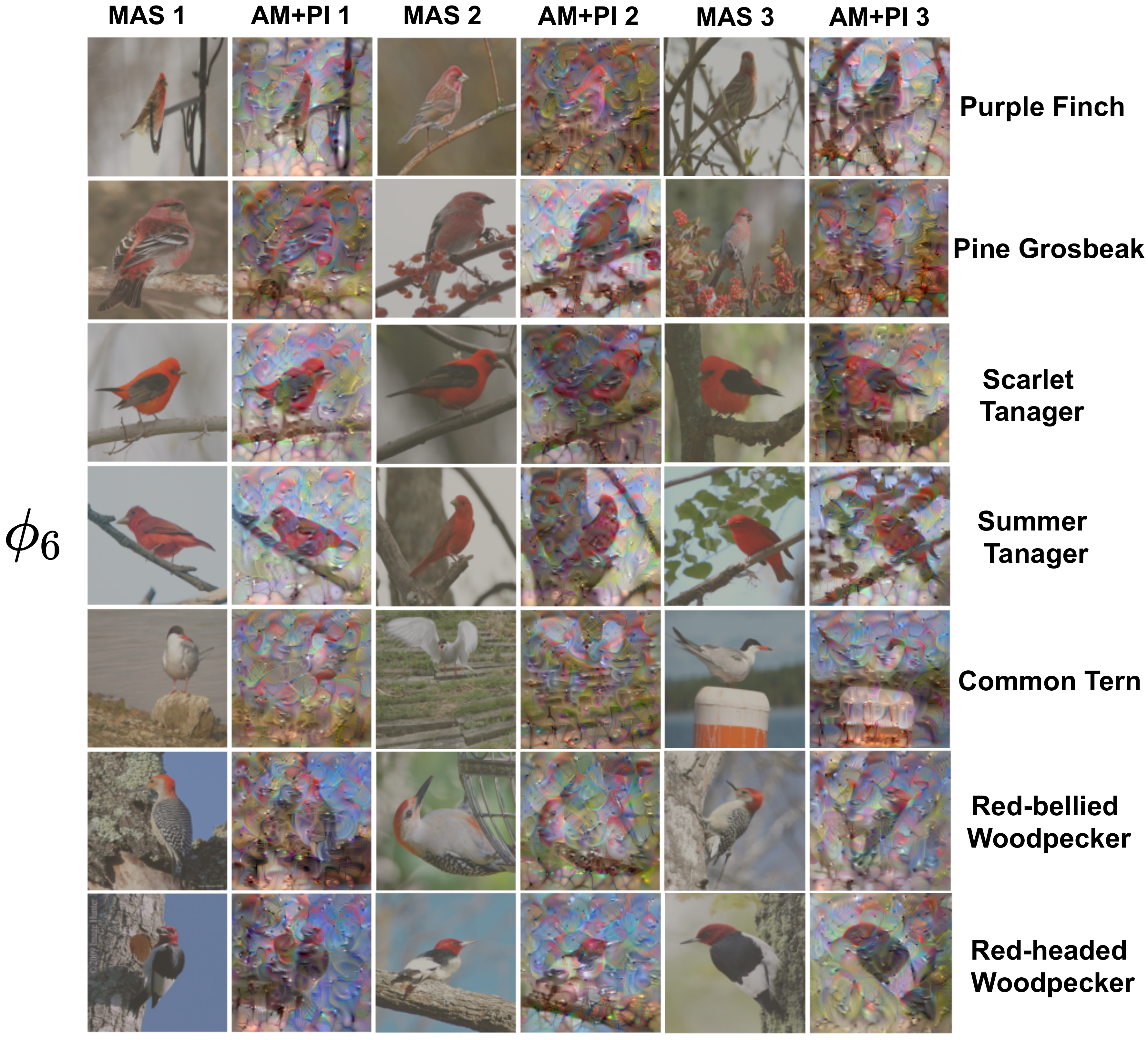}
    \caption{Relevant class-attribute pairs on CUB-200 with attribute $\phi_{6}$. Each row gives visualization for a relevant class of the attribute with three MAS and corresponding AM+PI outputs.}
    \label{cub200_global_2}
\end{figure}

\begin{figure}[!htb]
    \centering
    \includegraphics[width=1.11\textwidth]{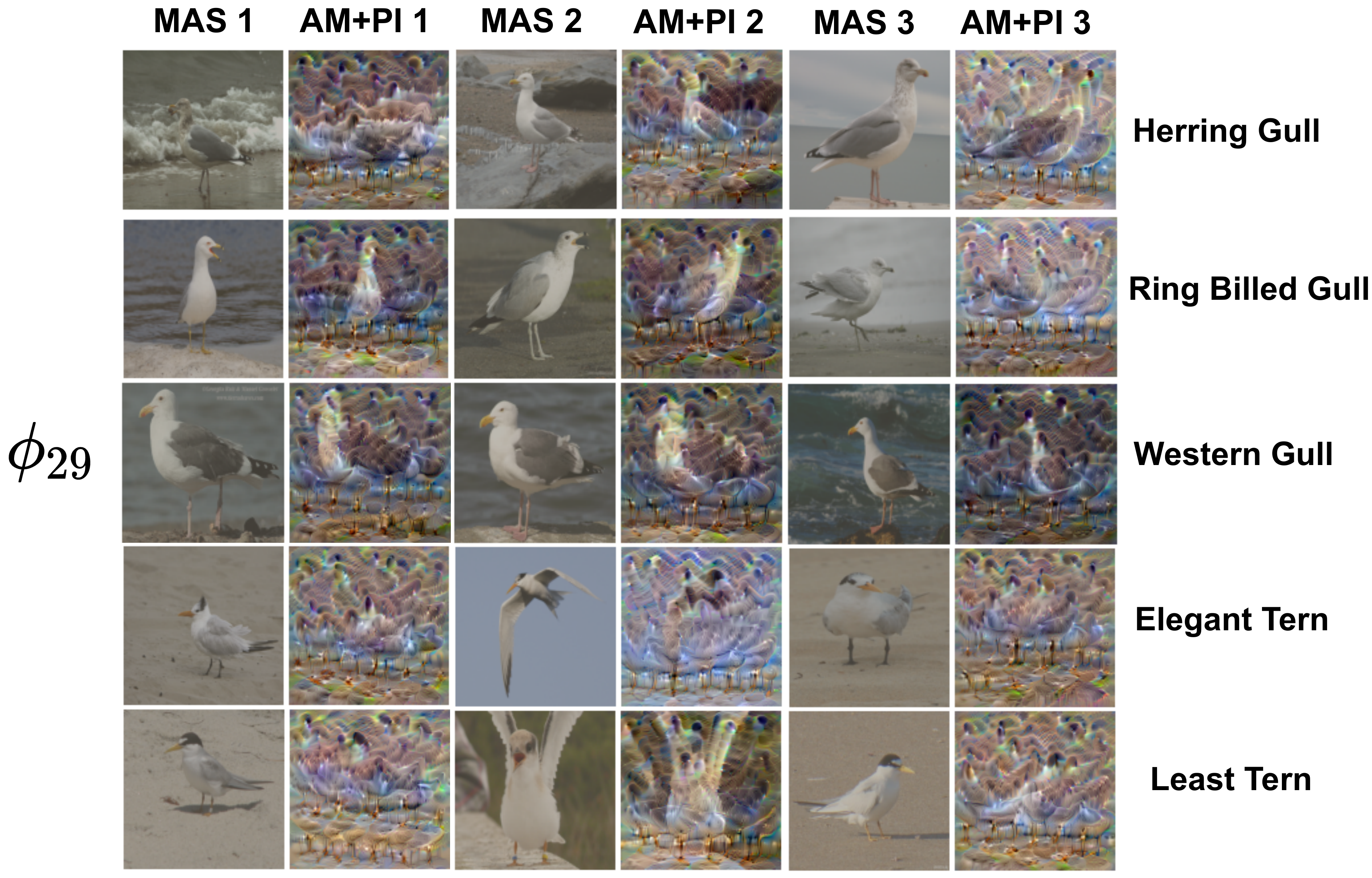}
    \caption{Relevant class-attribute pairs on CUB-200 with attribute $\phi_{29}$. Each row gives visualization for a relevant class of the attribute with three MAS and corresponding AM+PI outputs.}
    \label{cub200_global_4}
\end{figure}

We report the accuracy of BASE-$f$, FLINT-$f$ and FLINT-$g$ models (single run) and fidelity of FLINT-$g$ to FLINT-$f$ in Tab. \ref{results_additional} and conciseness below in Fig. \ref{cns_additional}. It should be noted that due to high number of classes, the disagreements between $f$ and $g$ are more common. The generated interpretations for the class predicted by $f$ can still be useful if it is among top classes predicted by $g$ (for a more detailed discussion, see Sec. \ref{disagree}). Thus we report below top-$k$ fidelity of $g$ to $f$ for $k=1, 5$ (the default fidelity of interpreter metric corresponds to $k=1$). We also illustrate visualizations of sample relevant class-attribute pairs with global relevance $r_{j, c}>0.5$ in Fig. \ref{cifar100_global_viz} for CIFAR100, and for CUB-200 in Figs. \ref{cub200_global_1}, \ref{cub200_global_2}, \ref{cub200_global_3} , \ref{cub200_global_4}.

\begin{table}
\small
\centering
\begin{tabular}{l c | c c c c c}
\toprule
\multicolumn{1}{c}{} &
\multicolumn{3}{c}{Accuracy (in \%)} & \multicolumn{2}{c}{Fidelity (in \%)} \\
\cmidrule[1pt](lr){2-4} 
\cmidrule[1pt](lr){5-6}
Dataset & BASE-$f$ & FLINT-$f$ & FLINT-$g$ & Top-1 & Top-5\\
\midrule
CIFAR100 & 70.7 & 70.8 & 69.9 & 85.2 & 97.3\\
CUB-200 & 71.3 & 71.0 & 68.7 & 80.0 & 96.7\\
\bottomrule
\vspace{1pt}
\end{tabular}
\caption[Results]{
Results for accuracy (in \%) and fidelity to FLINT-$f$ on CIFAR100, CUB-200.
}
\label{results_additional}
\end{table}

\textbf{Key observations}: FLINT-$f$ achieves almost the same accuracy as BASE-$f$ model for both datasets, competitive for models of this size. Given the large number of classes, it achieves high fidelity of interpretations with top-1 fidelity of more than 80\% and top-5 fidelity around 97\% for both datasets. The effect of increased number of classes and complexity of datasets is also seen in comparatively higher conciseness of FLINT. However, relative to the total number of attributes, the interpretations still utilize small fraction of them, similar to results on other datasets. We also showcase the ability of attributes learnt in FLINT to capture interesting concepts. For eg. on CUB-200, we visualize various attributes which encode concepts like 'yellow-headed birds' (Fig. \ref{cub200_global_1}), 'red-headed birds' (Fig. \ref{cub200_global_2}), 'blue-faced birds' (Fig. \ref{cub200_global_3}) and 'long orange/red legs' (Fig. \ref{cub200_global_4}).    





\begin{figure}[!htb]
    \centering
    \includegraphics[width=0.48\textwidth]{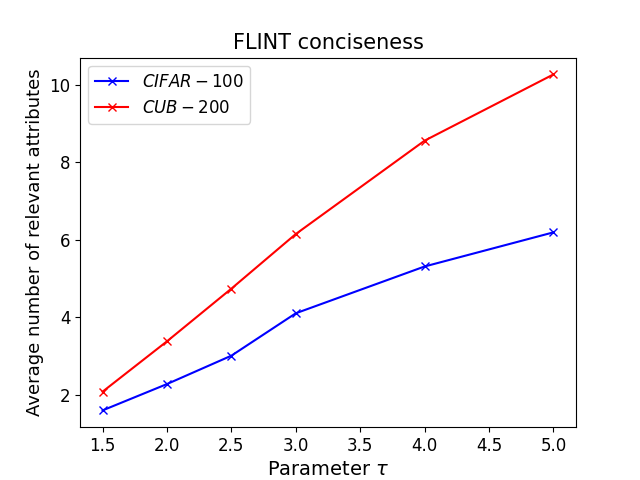}
    \caption{Conciseness curve of FLINT-$g$ interpretations on CIFAR100 and CUB-200}
    \label{cns_additional}
\end{figure}

\vspace{9pt}

\subsection{Ablation Studies and Analysis}
\label{analysis_viz}

\subsubsection{Shuffling experiment}

By structure, for both FLINT-$g$ and SENN, the output are generated by combining high level attributes and weights. To test how crucial the learnt attributes are to their predictions, we shuffle the attribute values $\Phi(x)$ for each sample $x$ (this corresponds to shuffling $h(x)$ for SENN with their notations). This is an extreme test: we therefore expect an important drop in accuracy. Tab. \ref{shuffle_1} reports the results for the experiments for our method and SENN. More precisely, we calculate the drop in prediction accuracy of FLINT-$g$ (and SENN), compared to their mean accuracies. For SENN, the very small drop in accuracy indicates its robustness to this shuffling, which highlights the fact that in this model, the activation of a given subset of attributes is not crucial for the prediction. In contrast FLINT-$g$ relies strongly on its attributes for its prediction.

\begin{table}[!hb]
\small
\centering
\begin{tabular}{l c c c} 
\toprule
Dataset & SENN & FLINT-$g$\\
\toprule
MNIST & 0.5 & 87.6\\
FashionMNIST & 10.9 & 76.6 \\ 
CIFAR-10 & 17.5 & 74.4\\
QuickDraw & 0.3 & 74.9\\
\bottomrule
\end{tabular}
\vspace{3pt}
\caption[Results]{FLINT and SENN accuracy drop for shuffled attributes (in \%)
}
\label{shuffle_1}
\end{table}

\subsubsection{Disagreeement analysis}
\label{disagree}
In this part, we analyse in detail the ``disagreement'' between the predictor $f$ and the interpreter $g$. Note that we already achieve very high fidelity to predictor for all datasets. We limit our analysis to QuickDraw, our dataset with least fidelity. Understanding disagreement can help us improving our framework as well as providing a measure of reliability about predictors output. 


For a given sample with disagreement, if the class predicted by $f$ is among the top predicted classes of $g$, the disagreement is acceptable to some extent as the attributes can still potentially interpret the prediction of $f$. The worse kind of samples for disagreement are the ones where class predicted by $f$ is not among the top predicted classes of $g$, and even worse are where, in addition to this, $f$ predicts the true label. We thus compute the top-$k$ fidelity (for $k=2, 3, 4$) on QuickDraw with ResNet, which for the default parameters described in the main paper, achieves a top-$2$ fidelity of $94.7\%$, top-$3$ fidelity $96.9\%$, and top-$4$ fidelity $98.2\%$. Only on $141$ (i.e. $0.7\%$) samples the class predicted by $f$, same as true class, is not in top-$3$ predicted by $g$ classes. 

\begin{figure}[t]
\centering
\includegraphics[width=0.4\textwidth]{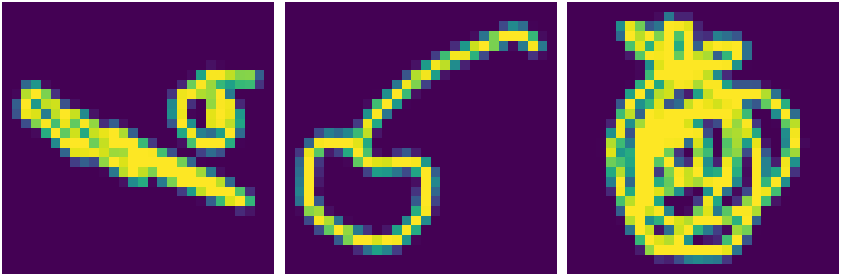} 
\caption{The three 'Apple' class samples classified correctly by $f$ but not by $g$.}
\label{disag_apple}
\end{figure}

\begin{figure}[t]
\centering
\includegraphics[width=0.46\textwidth]{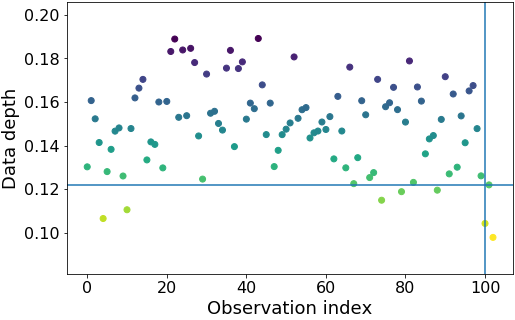} 
\caption{Projection data depth calculated with \eqref{prjdepth} w.r.t. the $8000$ 'Apple' training sample for $100$ 'Apple' test samples and for the three (observation indices $101$--$103$) 'Apple' class samples classified correctly by $f$ but not by $g$.}
\label{depth_apple}
\end{figure}

For eg., for the 'Apple' class (in QuickDraw), there only three disagreement samples for which $f$ delivers correct prediction (plotted in Fig.~\ref{disag_apple}) are not resembling apples at all. We propose an original analysis approach that consists in calculating a {\it robust centrality measure}---the projection depth---of these three samples as well as of another $100$ training samples w.r.t. the $8000$ training 'Apple' samples, plotted in Fig.~\ref{depth_apple}. To that purpose, we use the notion of projection depth \citep{ZuoS00, Mosler13}  for a sample $\boldsymbol{x}\in\mathbb{R}^d$ w.r.t. a dataset $\boldsymbol{X}$ which is defined as follows:
\begin{equation}\label{prjdepth}
    D(\boldsymbol{x}|\boldsymbol{X}) = 
         \left(1 + \sup_{\boldsymbol{p}\in \mathcal{S}^{d-1}}
         \frac {|\langle \boldsymbol{p},\boldsymbol{x} \rangle
                - \mbox{med}(\langle \boldsymbol{p},\boldsymbol{X} \rangle)|}
               {\mbox{MAD}(\langle \boldsymbol{p},\boldsymbol{X} \rangle)}
         \right)^{-1} \,,
\end{equation}
with $\langle \cdot,\cdot \rangle$ denoting scalar product (and thus $\langle \boldsymbol{p},\boldsymbol{X} \rangle$ being a vector of projection of $\boldsymbol{X}$ on $\boldsymbol{p}$) and $\mbox{med}$ and $\mbox{MAD}$ being the univariate median and the median absolute deviation form the median.  Fig.~\ref{depth_apple} confirms the visual impression that these $3$ disagreement samples are outliers (since their depth in the training class is low).

\begin{figure}[t]
\centering
\includegraphics[width=0.6\textwidth]{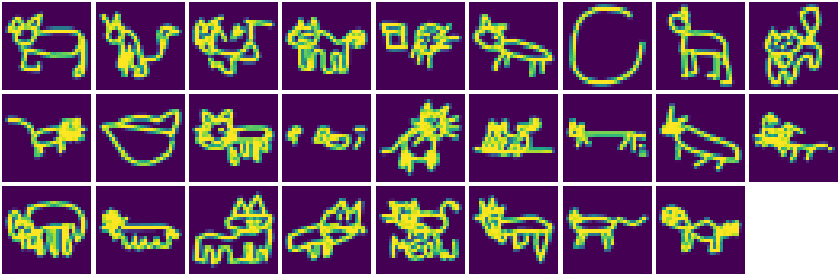} 
\caption{$26$ samples from 'Cat' class which are not in top$3$ $f$-predicted classes.}
\label{disag_cat}
\end{figure}

Fig.~\ref{disag_cat} depicts $26$ such cases for 'Cat' class to illustrate their logical dissimilarity. Being a complex model, the ResNet-based predictor $f$ still manages to learn to distinguish these cases (while $g$ does not), but in a way $g$ does not manage at all to explain.
Eventually, exploiting disagreement of $f$ and $g$ could be used as a means to measure trustworthiness. Deepening this issue is left for future works.

\subsubsection{Effect of autoencoder loss} 
\begin{figure}[ht]
\centering
 \includegraphics[width=0.5\textwidth]{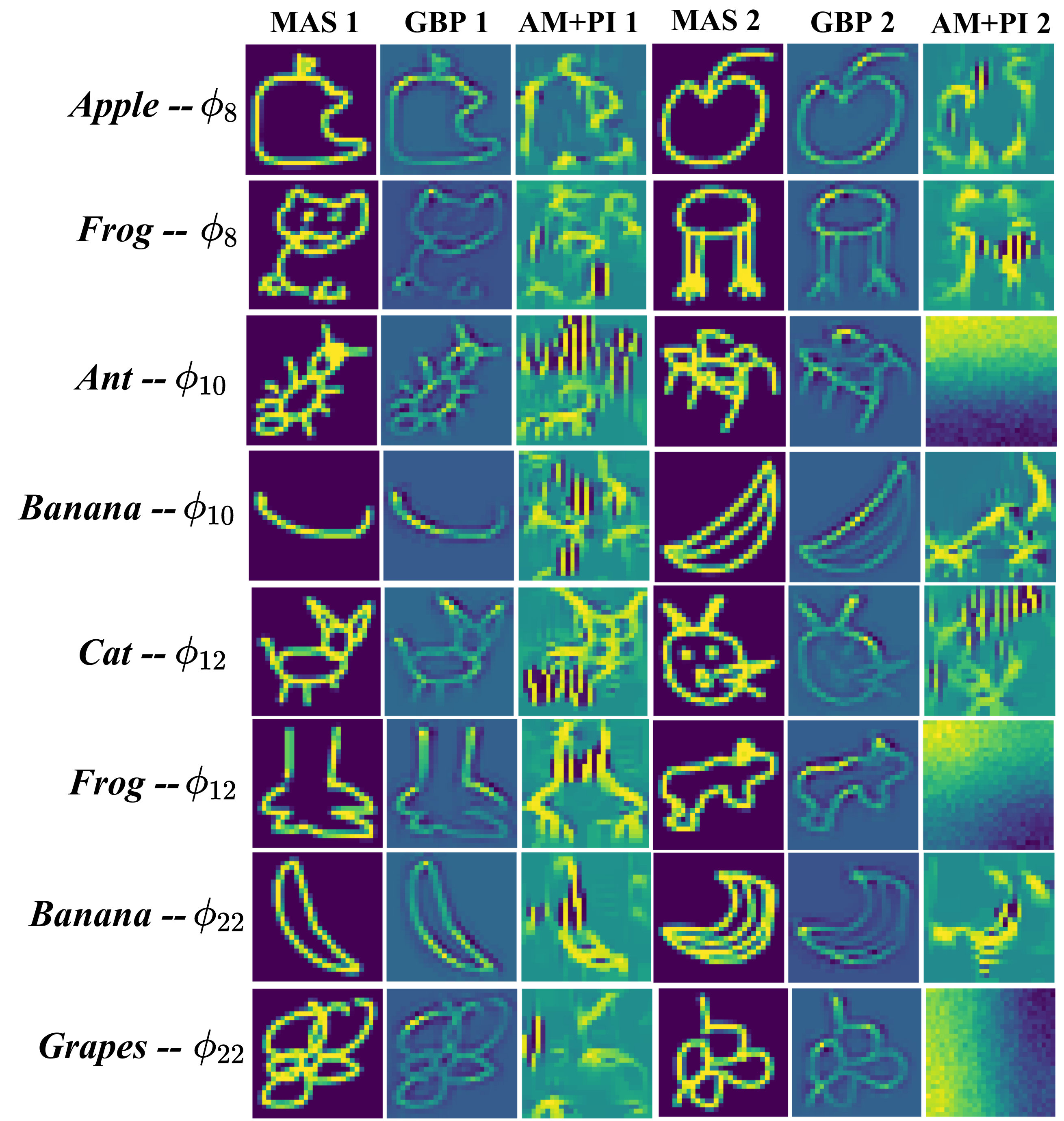}
\caption{Sample class-attribute pair visualizations learnt without autoencoder loss $\mathcal{L}_{if}$. GBP stands for Guided Backpropagation.}
\label{ae_1}
\end{figure}
Although the effect of $\mathcal{L}_{of}, \mathcal{L}_{cd}$ can be objectively assessed to some extent, the effect of $\mathcal{L}_{if}$ can only be seen subjectively. If the model is trained with $\gamma=0$, the attributes still demonstrate high overlap, nice conciseness. However, it becomes much harder to understand concepts encoded by them. For majority of attributes, MAS and the outputs of the analysis tools do not show any consistency of detected pattern. Some such attributes are depicted in Fig. \ref{ae_1} Such attributes are present even for the model trained with autoencoder, but are very few. We thus believe that autoencoder loss enforces a consistency in detected patterns for attributes. It does not necessarily guarantee semantic meaningfulness in attributes, however it's still beneficial for improving their understandability.

\subsubsection{Effect of hidden layer selection}

We already discussed the empirical rationale behind our choice of hidden layers in Sec. \ref{net_archs}. In general for any predictor architecture or dataset, the most obvious choice is to select last convolutional layer output. This also helps achieving high fidelity for $g$. The only problem that might arise when selecting layer(s) very close to the output is that the attribute might be learnt trivially. This is indicated by extremely low entropy and high input fidelity loss. While tuning hyperparameters of interpretability loss could be helpful in tackling this issue (reducing $\beta$, increasing $\gamma$), choosing an earlier hidden layer can also prove to be very useful. We study the effect of choice of hidden layers with ResNet18 on QuickDraw. We make 3 different choices of single hidden layers (9th, 13th, 16th conv layers). For each choice we tabulate resulting metrics (accuracy, fidelity of interpreter, reconstruction loss, conciseness for threshold $1/\tau=0.2$) in Tab. \ref{layer_effect}. All other hyperparameters remain same.

\begin{table}[ht]
\small
\centering
\begin{tabular}{l c c c c} 
\toprule
Layer & Accuracy (in \%) & Fidelity (in \%) & $\mathcal{L}_{if}$ & Conciseness $1/\tau=0.2$\\
\toprule

9th conv & 85.2 & 78.0 & 0.074 & 1.873\\
13th conv & 85.6 & 85.6 & \textbf{0.073} & 1.905\\
16th conv & 86.5 & \textbf{96.0} & 0.081 & \textbf{1.562}\\

\bottomrule
\end{tabular}
\vspace{3pt}
\caption[Results]{Effect of different hidden layers for Resnet18 on QuickDraw.}
\label{layer_effect}
\end{table}

\textbf{Key observations}: (a) Compared to average BASE-$f$ accuracy of 85.3\% for ResNet18 on QuickDraw, accuracy of all models are comparable or slightly better. Thus, choice of hidden layers does not strongly affect predictor accuracy. (b) The interpreter fidelity gets considerably better if the layer chosen is closer to the output. (c) The input fidelity/reconstruction loss does not behave as monotonously, but it is not surprising that layers close to the output result in worse input reconstruction. (d) Interpretations are expected to be more concise when chosen layer is very close to the output in the sense that conciseness is an indicator of abstraction level of the interpretation. Thus, a standard choice is to start with a layer close to the output. A small revision may be needed depending upon optimization of input fidelity loss.

\subsubsection{Effect of number of attributes J}

\begin{figure}[t]
\centering
\includegraphics[width=0.5\textwidth]{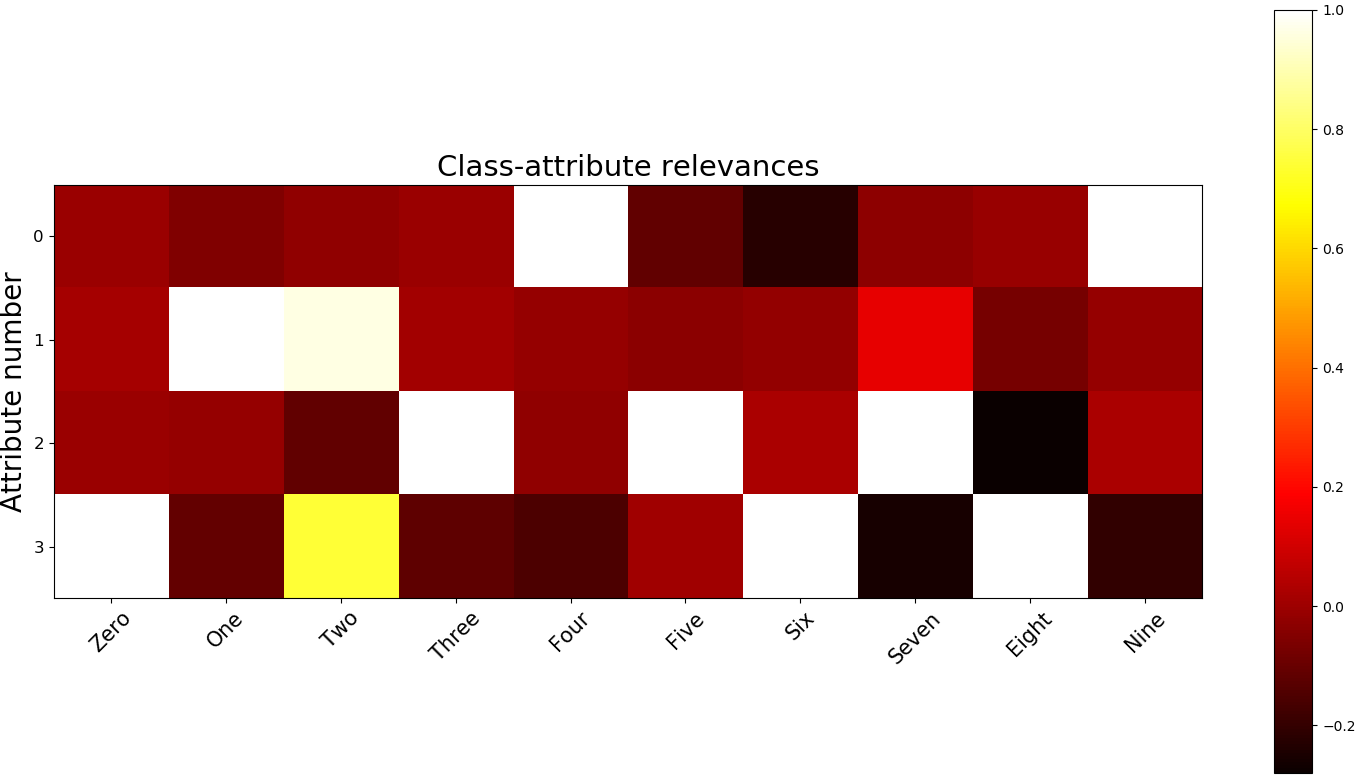} 
\caption{Global class attribute relevances for model with $J=4$ on MNIST.}
\label{J4_cac}
\end{figure}

\begin{figure}[t]
\centering
\includegraphics[width=0.6\textwidth]{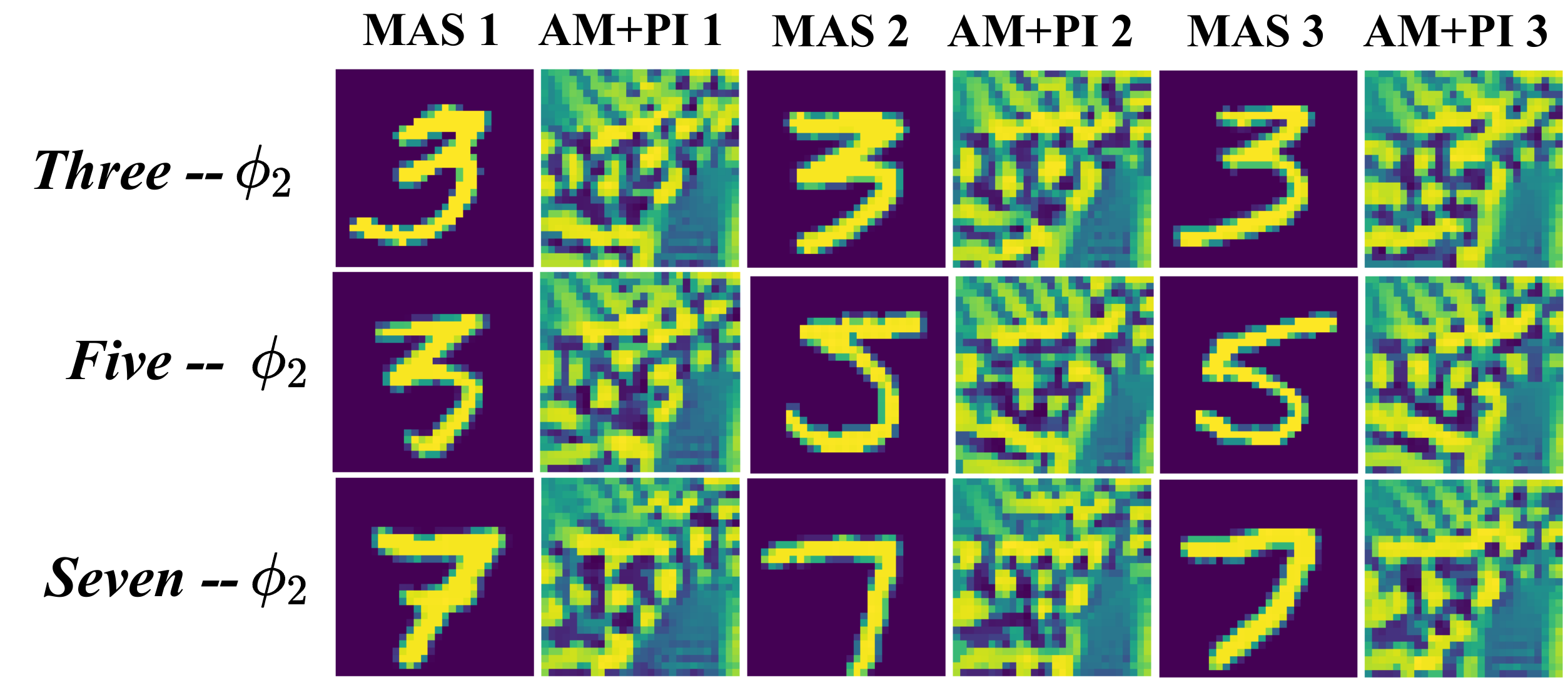} 
\caption{Interpretation for attribute $\phi_2$ for model learn on MNIST with $J=4$.}
\label{J4_int}
\end{figure}

\paragraph{Effect of $J$} We study the effect of choosing small values for number of attributes $J$ (keeping all other hyperparameters same). Tab. \ref{result_J_mnist} tabulates the values of input fidelity loss $\mathcal{L}_{if}$, output fidelity loss $\mathcal{L}_{of}$ on the training data by the end of training for MNIST and the fidelity of $g$ to $f$ on MNIST test data for different $J$ values. Tab. \ref{result_J_qd} tabulates same values for QuickDraw. The two tables clearly show that using small $J$ can harm the autoencoder and the fidelity of interpreter. Moreover, the system packs more information in each attribute and this makes it hard to understand them, specially for very small $J$. This is illustrated in Figs.~\ref{J4_cac} and \ref{J4_int}, which depict part of global interpretations generated on MNIST for $J=4$ (all the parameters take default values). Fig.~\ref{J4_cac} shows global class-attribute relevances and Fig.~\ref{J4_int} shows generated interpretation for a sample attribute $\phi_2$. It can be clearly seen that the attributes start encoding concepts for too many classes (high number of bright spots). This also causes their AM+PI outputs to be muddled with two many patterns. This adds a lot of difficulty in understandability of these attributes.

\begin{table}
\centering
\begin{tabular}{l c c c} 
\toprule
 & $\mathcal{L}_{if}$ (train) & $\mathcal{L}_{of}$ (train) & Fidelity (test) (\%)\\
\toprule
$J=4$ & 0.058 & 0.57 & 87.4\\
\midrule
$J=8$ & 0.053 & 0.23 & 97.5\\
\midrule
$J=25$ & 0.029 & 0.16 & 98.8\\
\bottomrule
\end{tabular}
\caption[Results]{Effect of $J$ on losses and fidelity for MNIST with LeNet.}
\label{result_J_mnist}
\end{table}

\begin{table}
\centering
\begin{tabular}{l c c c} 
\toprule
 & $\mathcal{L}_{if}$ (train) & $\mathcal{L}_{of}$ (train) & Fidelity (test) (\%)\\
\toprule
$J=4$ & 0.094 & 2.08 & 19.5\\
\midrule
$J=8$ & 0.079 & 1.48 & 57.6\\
\midrule
$J=24$ & 0.069 & 0.34 & 90.8\\
\bottomrule
\end{tabular}
\caption[Results]{Effect of $J$ on losses and fidelity for QuickDraw with ResNet.}
\label{result_J_qd}
\end{table}

\paragraph{How to choose the number of attributes} Assuming a suitable architecture for decoder $d$, simply tracking $\mathcal{L}_{if}, \mathcal{L}_{of}$ on training data can help rule out very small values of $J$ as they result in poorly trained decoder and relatively poor fidelity of $g$. One can also qualitatively analyze the generated explanations from the training data to tune $J$ to a certain extent. Too small values of $J$ can result in attributes encoding concepts for too many classes, which affects negatively their understandability. It is more tricky and subjective to tune $J$ once it becomes large enough so that $\mathcal{L}_{if}, \mathcal{L}_{of}$ are optimized well. The upper threshold of choosing $J$ is subjective and highly affected by how many attributes the user can keep a tab on or what fidelity user considers reasonable enough. It is possible that due to enforcement of conciseness, even for high value of $J$, only a small subset of attributes are relevant for interpretations. Nevertheless, for high $J$ value, there is a risk of ending up with too many attributes or class-attribute pairs to analyze.

It is important to  notice that it is possible to select $J$ from the training set only by using a cross-validation strategy. In practise, it seems reasonable to agree on smallest value of $J$ for which the  increase of the cross-validation fidelity estimate drops dramatically, since further increase of $J$ would generate less understandable attributes with very little gain in fidelity.

\subsubsection{Effect of loss scheduling}

We also study the effect of introduction of different schedules for output fidelity and conciseness loss with ResNet18 on CIFAR10. We introduce $\bmL_{of}$ at different points of time during training (indicated by first column of Tab. \ref{loss_effect}. $\bmL_{cd}$ is introduced 1 epoch later. The first row corresponds to current setting proposed in the main paper. Total training time constitutes 25 epochs. All other hyperparameters remain same.

\begin{table}[ht]
\small
\centering
\begin{tabular}{l c c c c} 
\toprule
Time of introduction & Accuracy (in \%) & Fidelity (in \%) & $\mathcal{L}_{if}$ & Conciseness $1/\tau=0.2$\\
\toprule

Epoch 3 (current) & 84.6 & 93.5 & 0.421 & 2.612\\
Epoch 4 & 84.8 & 93.4 & 0.427 & 2.501\\
Epoch 5 & 84.3 & 94.2 & 0.426 & 2.351\\
Epoch 6 & 85.0 & 93.1 & 0.426 & 2.376\\
Epoch 8 & 84.5 & 93.7 & 0.432 & 2.642\\
Epoch 10 & 84.6 & 93.9 & 0.422 & 1.944\\
Epoch 14 & 84.2 & 92.1 & 0.445 & 2.274\\
Epoch 21 & 84.6 & 91.2 & 0.450 & 3.710\\
Epoch 24 & 84.4 & 86.3 & 0.524 & 4.533\\

\bottomrule
\end{tabular}
\vspace{3pt}
\caption[Results]{Effect of loss scheduling for Resnet18 on CIFAR10.}
\label{loss_effect}
\end{table}

\textbf{Key Observations}: (a) As soon as the system receives reasonable time to train with all three losses (note that input fidelity loss is always present), small changes to introduction of losses have little to no impact on the metrics. (b) By contrast, when we introduce the losses extremely late (for eg. see the last two rows), the interpretability losses/metrics get noticeably worse.

\subsubsection{Additional visualizations}
\label{additional_viz}

\begin{figure}[t!]
\centering
\includegraphics[width=0.7\textwidth]{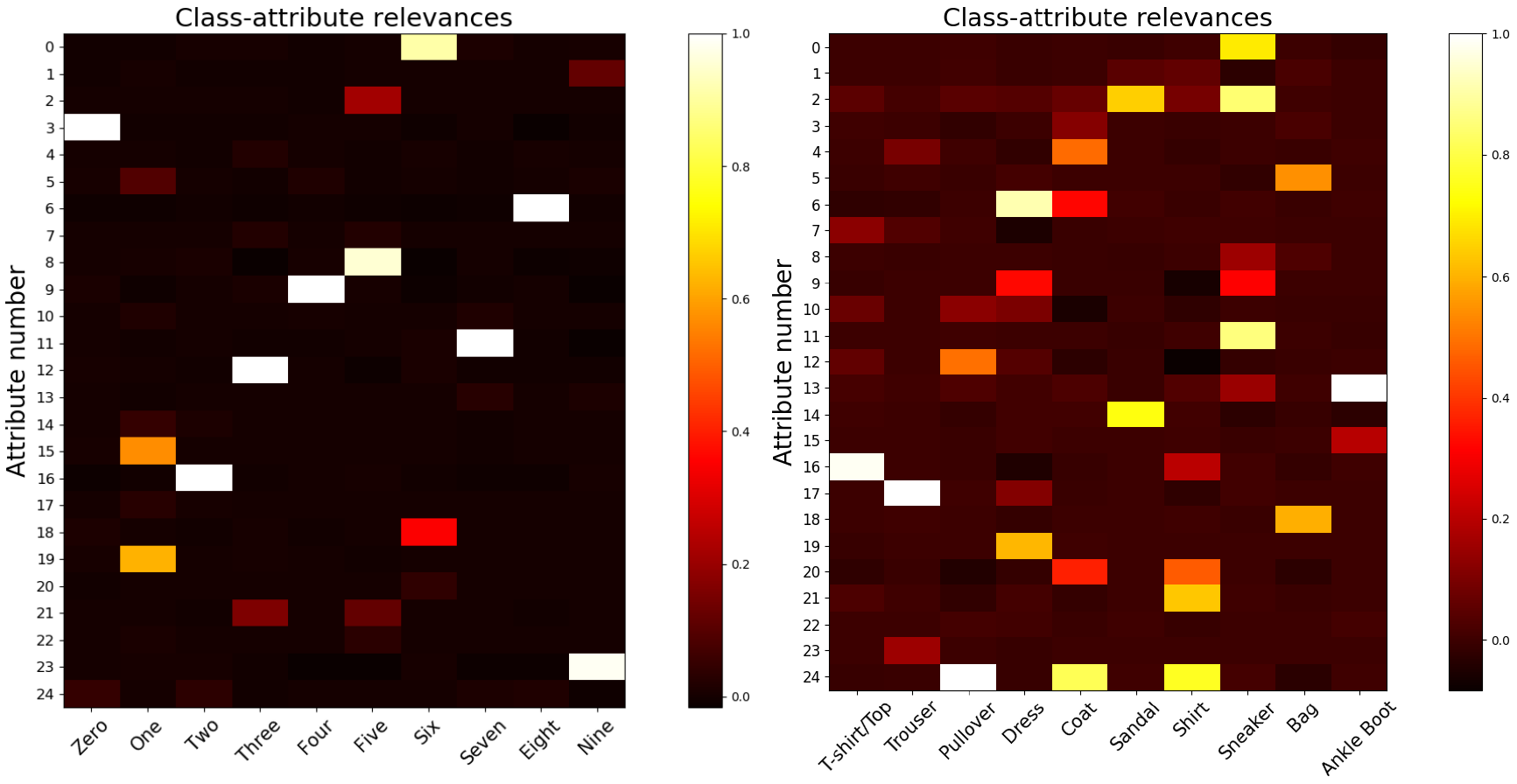}
\caption{Global class-attribute relevances $r_{j,c}$ for MNIST (Left) and FashionMNIST (Right). 14 class-attribute pairs for MNIST and 26 pairs for FashionMNIST have relevance $r_{j,c} > 0.2$.}
\label{global_rel_sup}
\end{figure}

\begin{figure}[t!]
\centering
\includegraphics[width=0.58\textwidth]{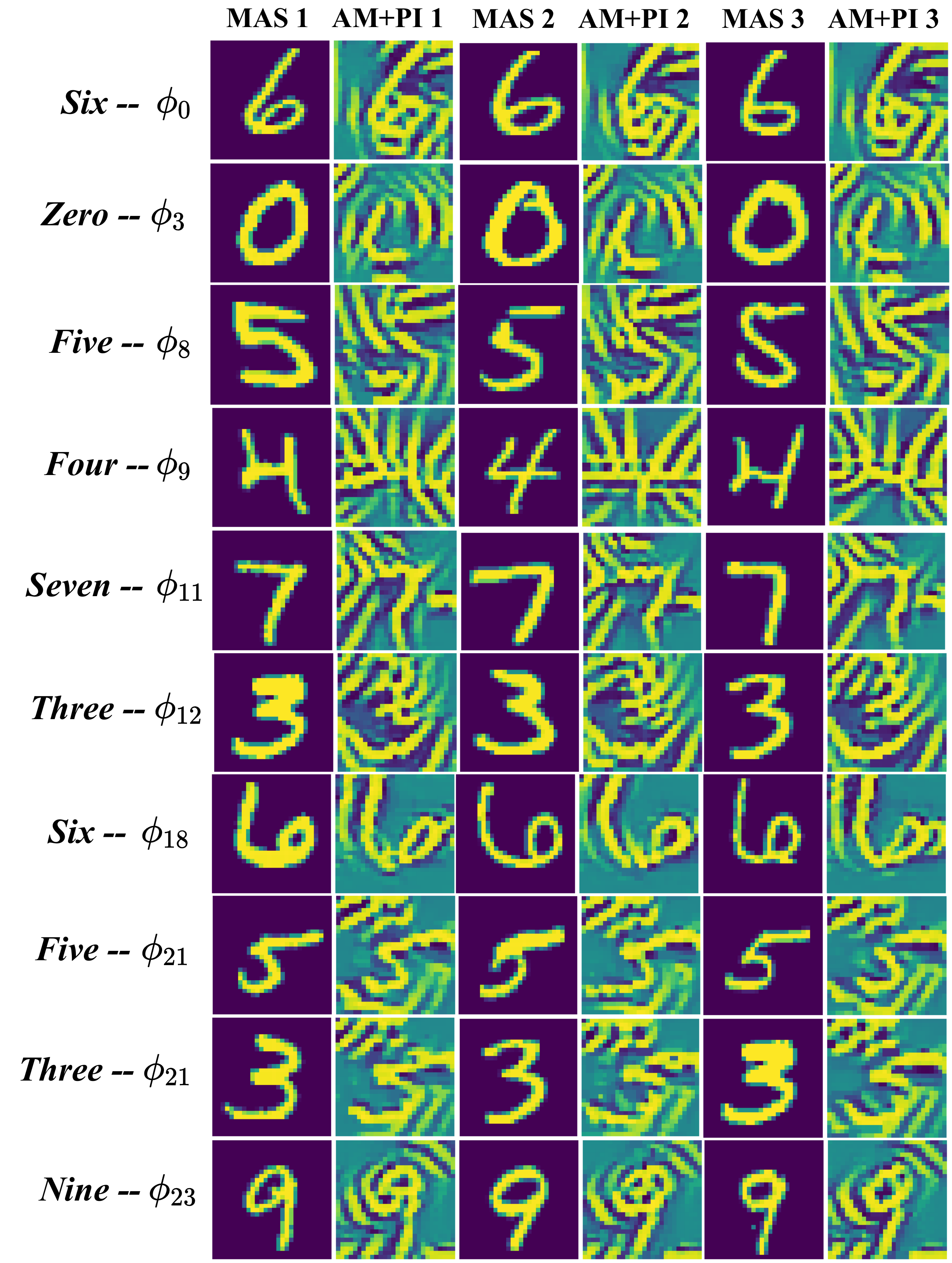} 
\caption{Additional class-attribute visualizations for MNIST. Three MAS and their corresponding AM+PI outputs are shown.}
\label{mnist_global_viz}
\end{figure}

\begin{figure}[t!]
\centering
\includegraphics[width=0.58\textwidth]{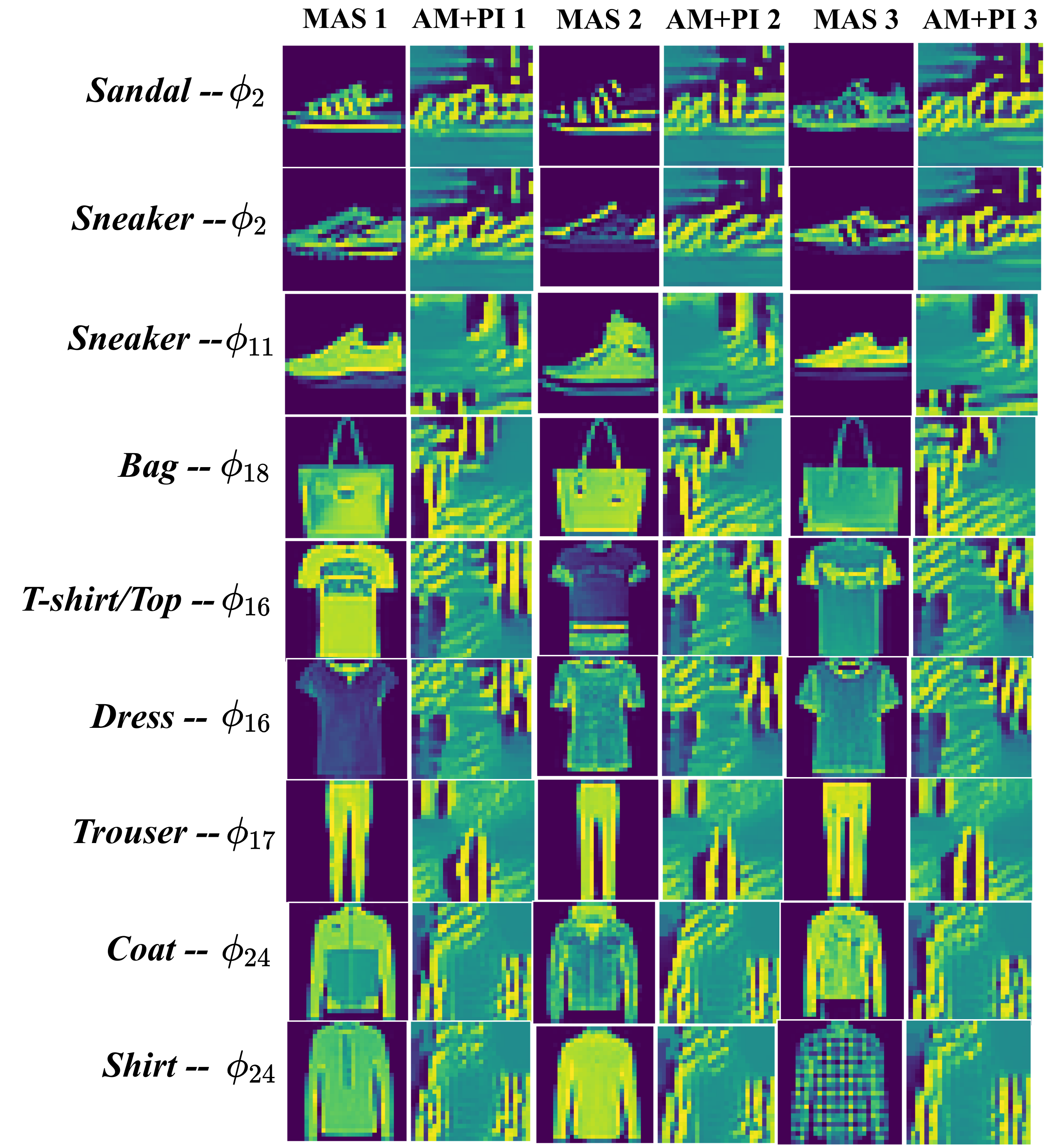} 
\caption{Additional class-attribute visualizations for Fashion-MNIST. Three MAS and their corresponding AM+PI outputs are shown.}
\label{fmnist_global_viz}
\end{figure}

\begin{figure}[t]
\centering
\includegraphics[width=0.6\textwidth]{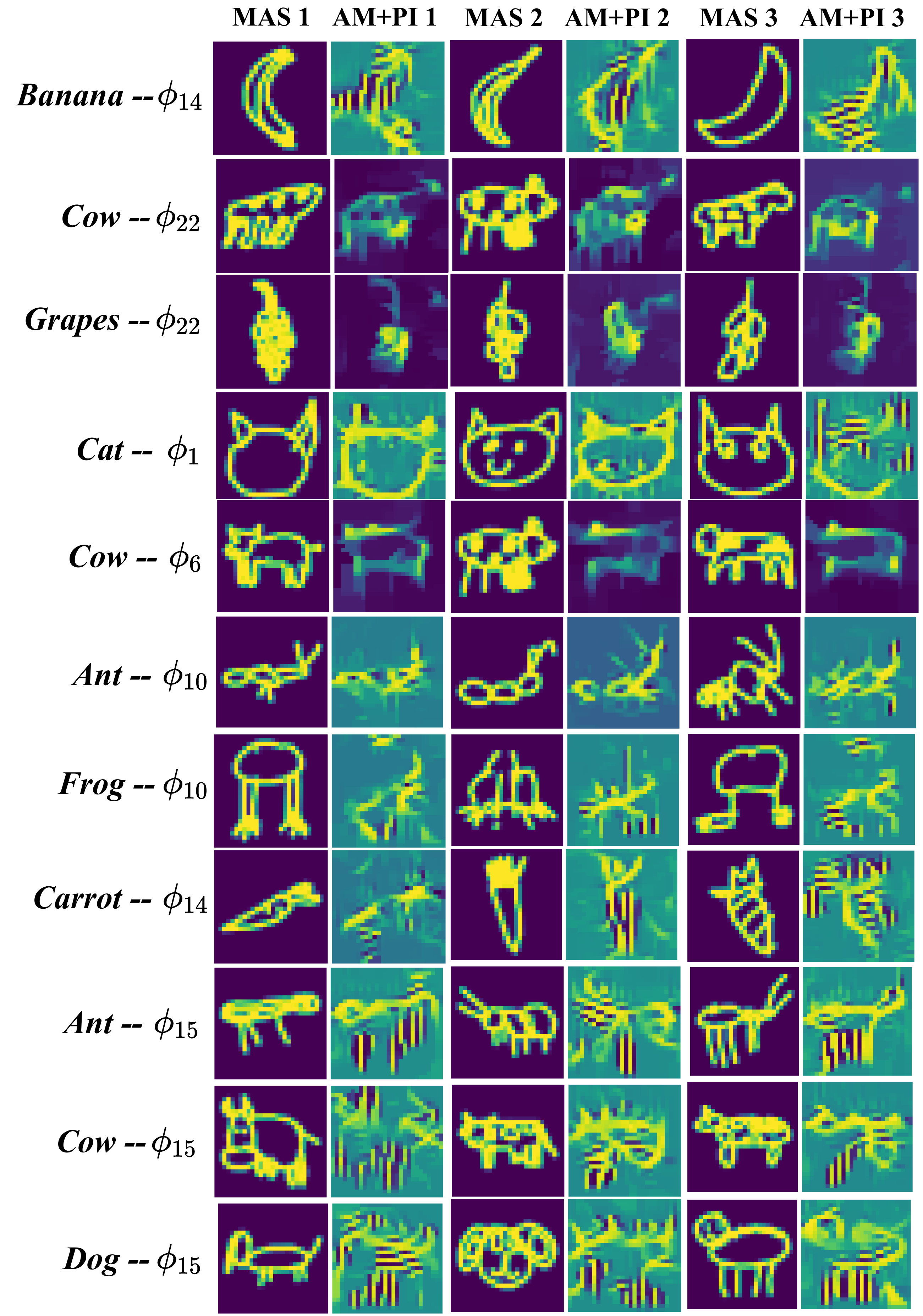} 
\caption{Additional class-attribute visualizations for QuickDraw. Three MAS and their corresponding AM+PI outputs are shown.}
\label{qdraw_global_viz}
\end{figure}

\begin{figure}[t!]
\centering
\includegraphics[width=0.6\textwidth]{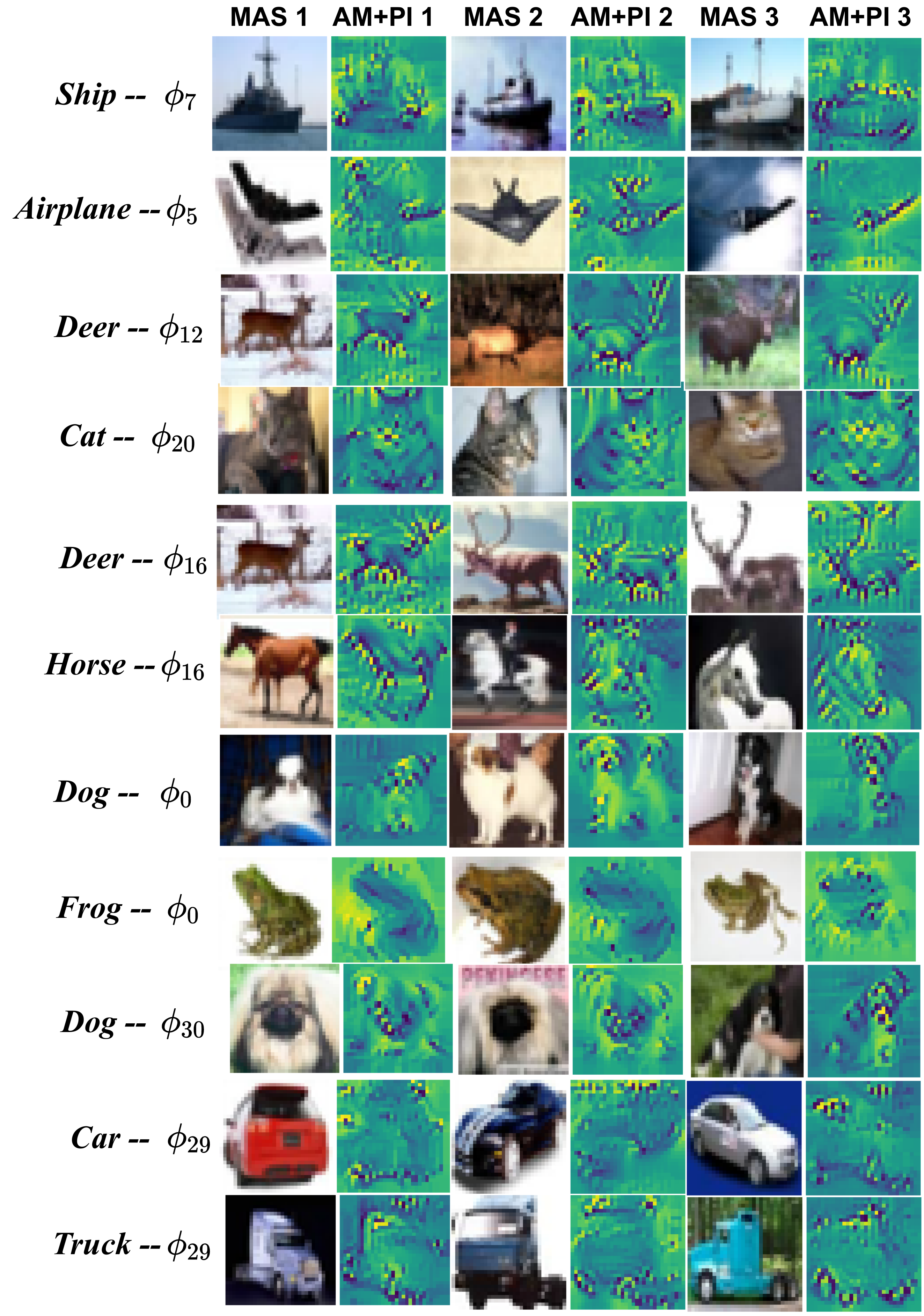} 
\caption{Additional class-attribute visualizations for CIFAR-10. Three MAS and their corresponding AM+PI outputs are shown.}
\label{cifar10_global_viz}
\end{figure}

\begin{figure}[t!]
\centering
\includegraphics[width=0.5\textwidth]{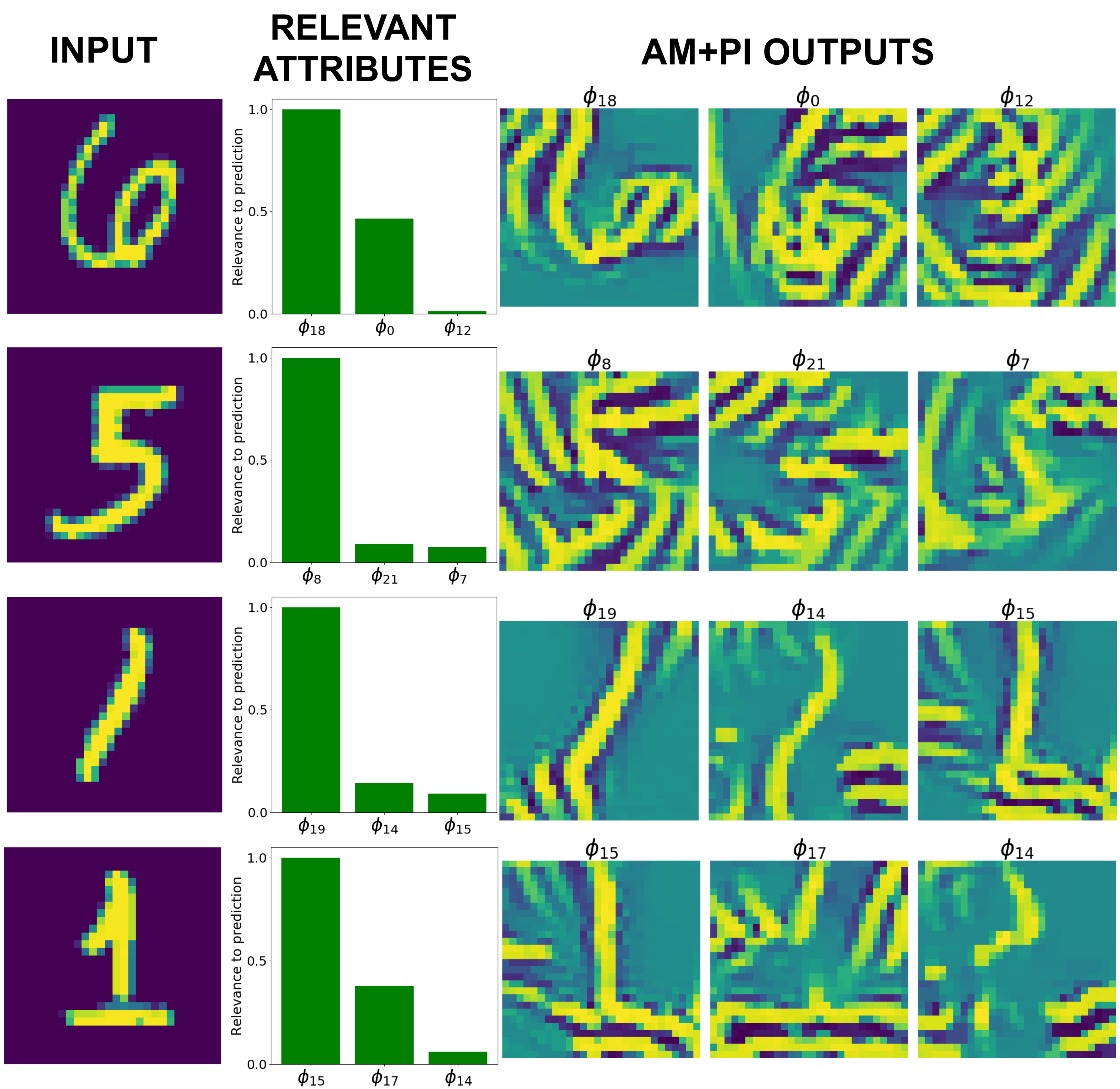} 
\caption{Local interpretations on test samples for MNIST. True labels are: 'Six', 'Five', 'One' and 'One'. Top 3 most relevant attributes and their corresponding AM+PI outputs are shown.}
\label{mnist_local_viz}
\end{figure}

\begin{figure}[t!]
\centering
\includegraphics[width=0.5\textwidth]{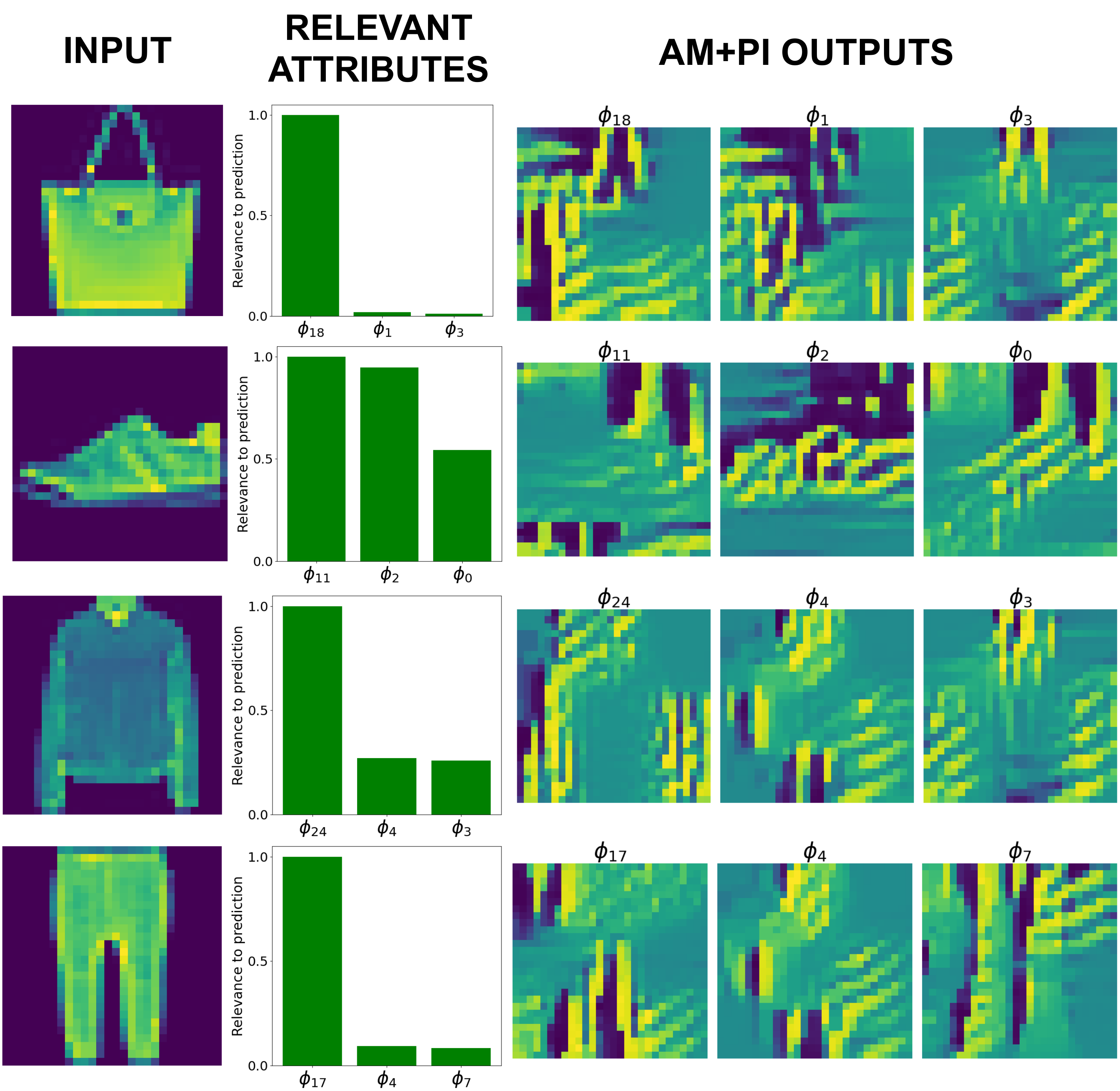} 
\caption{Local interpretations on test samples for Fashion-MNIST. True labels are: 'Bag', 'Sneaker, 'Coat', 'Trousers'. Top 3 most relevant attributes and their corresponding AM+PI outputs are shown.}
\label{fmnist_local_viz}
\end{figure}

\begin{figure}[t!]
\centering
\includegraphics[width=0.5\textwidth]{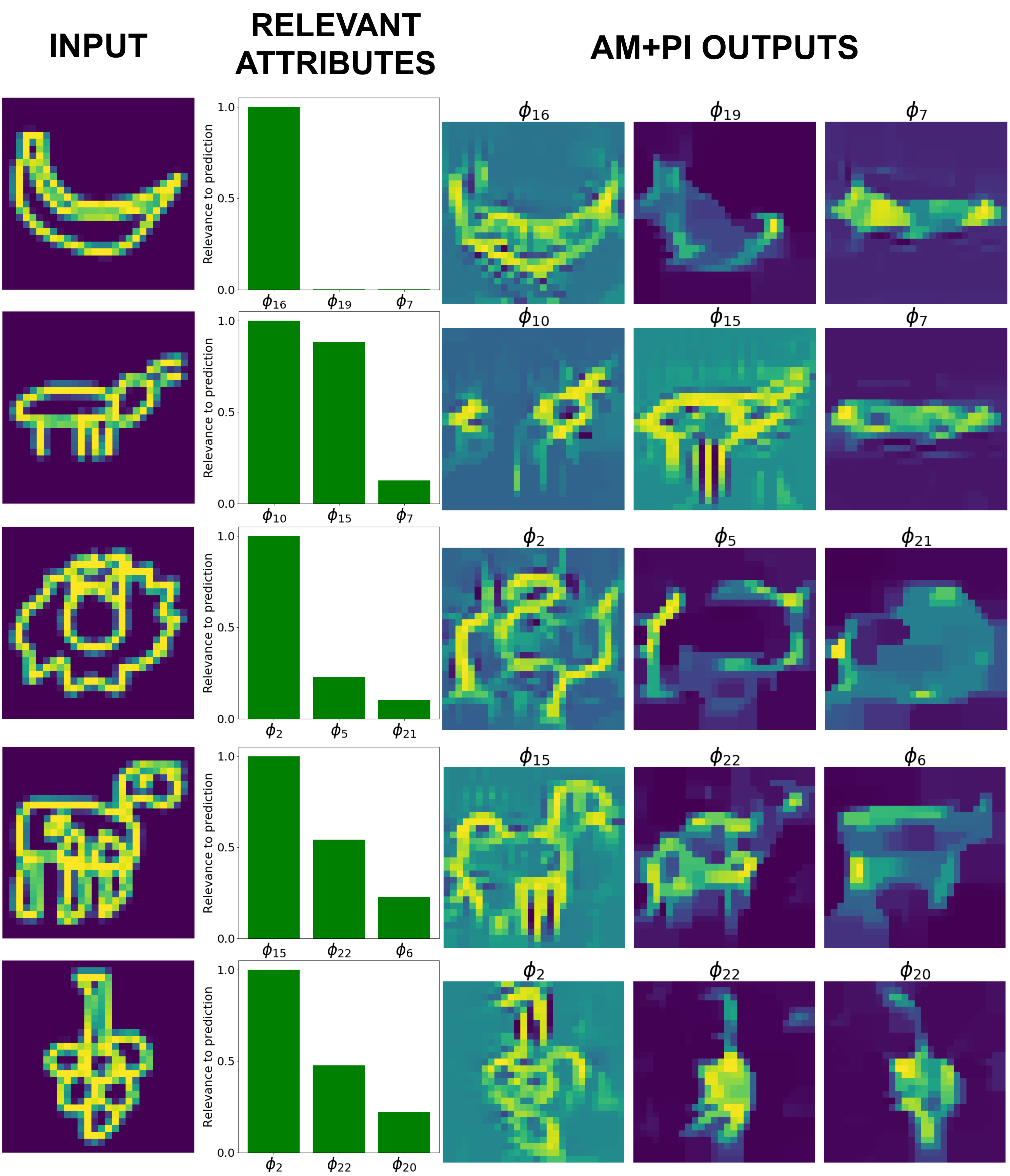} 
\caption{Local interpretations on test samples for QuickDraw. True labels are: 'Banana', 'Ant', 'Lion', 'Cow' and 'Grapes'. Top 3 most relevant attributes and their corresponding AM+PI outputs are shown.}
\label{qdraw_local_viz}
\end{figure}

\begin{figure}[ht!]
\centering
\includegraphics[width=0.5\textwidth]{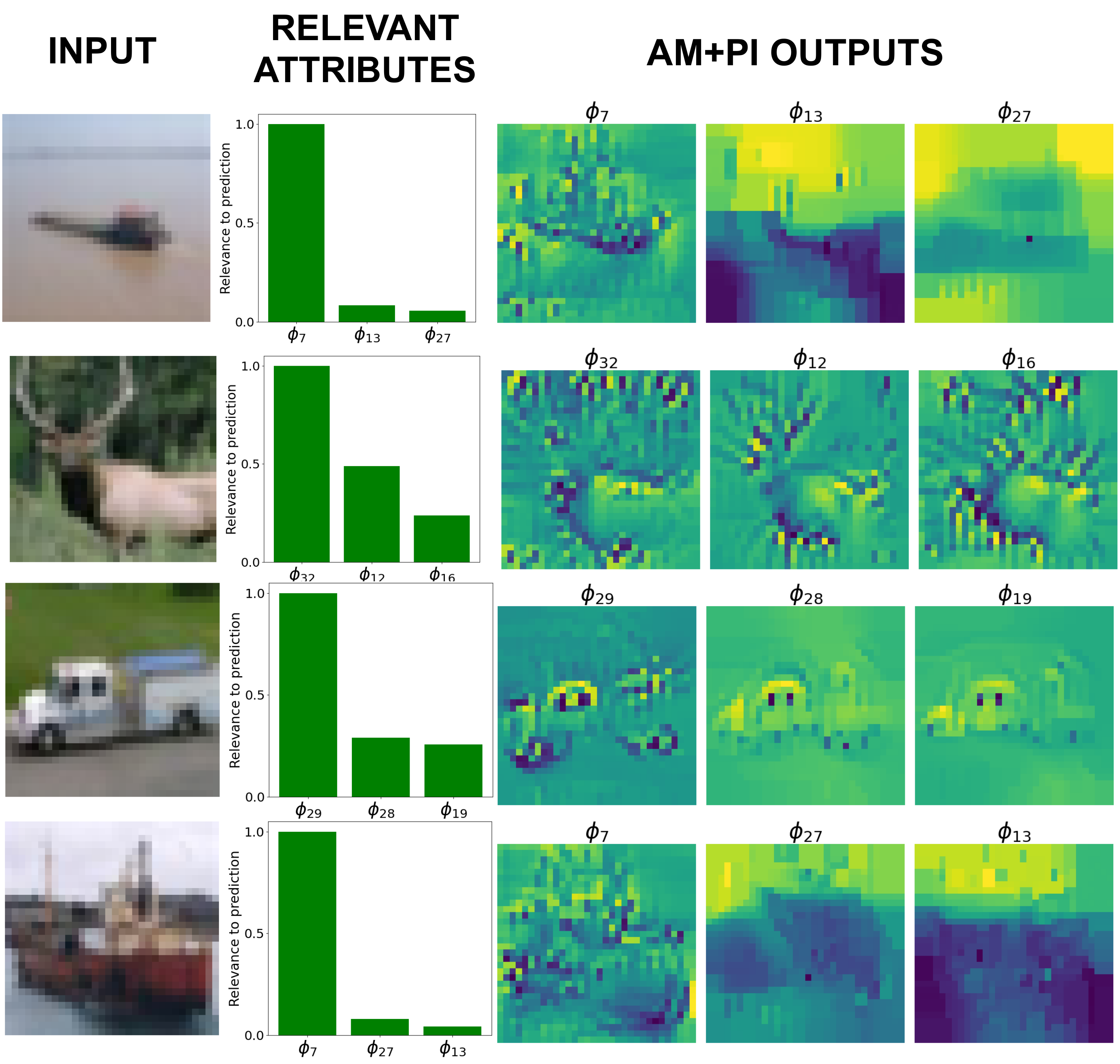} 
\caption{Local interpretations on test samples for CIFAR-10. True labels are: 'Ship', 'Deer', 'Truck' and 'Ship'. Top 3 most relevant attributes and their corresponding AM+PI outputs are shown.}
\label{cifar10_local_viz}
\end{figure}

For completeness, we show some additional visualizations of global interpretations (relevances, class-attribute pairs) and local interpretations.

Fig. \ref{global_rel_sup} contains global relevances generated for MNIST and FashionMNIST. Global relevances for QuickDraw and CIFAR10 are in main paper.

Figs. \ref{mnist_global_viz}, \ref{fmnist_global_viz}, \ref{qdraw_global_viz}, \ref{cifar10_global_viz} show some additional class-attribute pairs and their visualizations for all 4 datasets. Local interpretations on some test samples from these datasets are depicted in Figs. \ref{mnist_local_viz}, \ref{fmnist_local_viz}, \ref{qdraw_local_viz}, \ref{cifar10_local_viz}.


\subsection{Other tools for analysis}
\label{other_tools}

\begin{figure*}[t]
\centering
\includegraphics[width=1.01\textwidth]{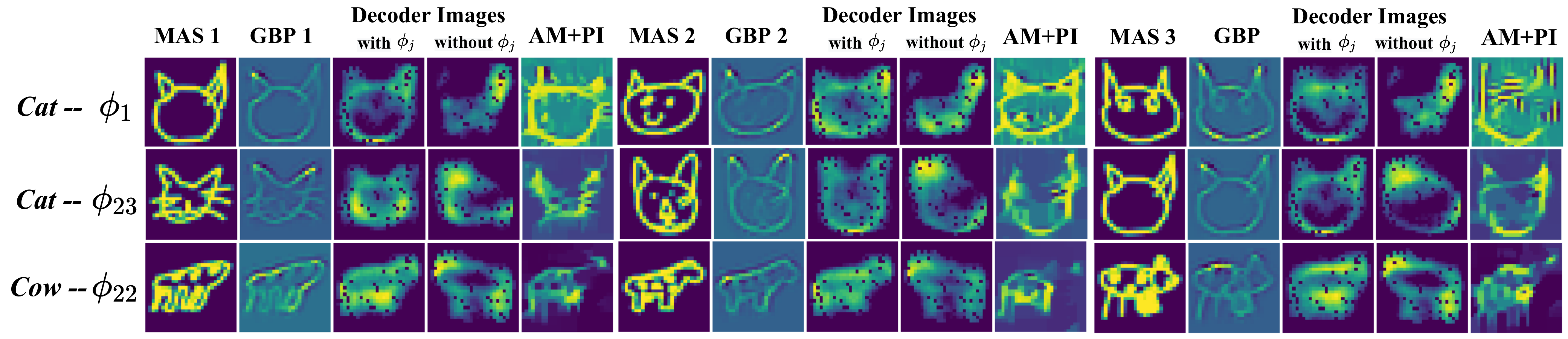} 
\caption{Examples of class-attribute pairs on QuickDraw, where decoder assists in understanding of encoded concept for the attribute.}
\label{ex_dec}
\end{figure*}

\begin{figure*}[t]
\centering
\includegraphics[width=1.01\textwidth]{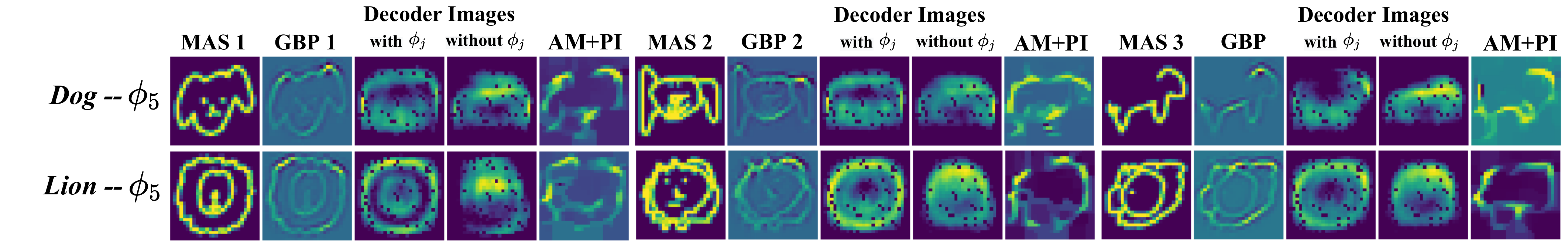} 
\caption{Examples of class-attribute pairs on QuickDraw, where input attribution (GBP) assists in understanding of encoded concept for the attribute. GBP stands for Guided Backpropagation.}
\label{ex_gbp}
\end{figure*}

Although we consider AM+PI as the primary tool for analyzing concepts encoded by attributes (for MAS of each class-attribute), other tools can also be helpful in deeper understanding of the attributes. We introduce two such tools:
\begin{itemize}
    \item \textit{Input attribution}: This is a natural choice to understand an attribute's action for a sample. Any algorithms ranging from black-box local explainers to saliency maps can be employed. These maps are less noisy (compared to AM+PI) and very general choice, applicable to almost all domains.
    
    \item \textit{Decoder}: Since we also train a decoder $d$ that uses the attributes as input. Thus, for an attribute $j$ and $x$, we can compare the reconstructed samples $d(\Phi(x))$ and $d(\Phi(x) \backslash j)$ where $\Phi(x) \backslash j$ denotes attribute vector with $\phi_j(x) = 0$, i.e., removing the effect of attribute $j$. While, the above comparison can be helpful in revealing information encoded in attribute $j$, it is not guaranteed to do so as the attributes can be entangled.

\end{itemize}

We illustrate the use of these tools for certain example class-attribute pairs on QuickDraw in Fig. \ref{ex_dec} and \ref{ex_gbp}. Note that as discussed in the main paper, these tools are not guaranteed to be always insightful, but their use can help in some cases. 

Fig. \ref{ex_dec} depicts example class-attribute pairs where decoder $d$ contributes in understanding of attributes. The with $\phi_j$ column denotes the reconstructed sample $d(\Phi(x))$ for the maximum activating sample $x$ under consideration. The without $\phi_j$ column is the reconstructed sample $d(\Phi(x)) \backslash j)$ with the effect of attribute $\phi_j$ removed for the sample under consideration ($\phi_j(x) = 0$). For eg. $\phi_1, \phi_{23}$, strongly relevant for Cat class, detect similar patterns, primarily related to the face and ears of a cat. The decoder images suggest that $\phi_1$ very likely is more responsible for detecting the left ear of cat and $\phi_{23}$, the right ear. Similarly analyzing decoder images for $\phi_{22}$ in the third row reveals that it is likely has a preference for detecting heads present towards the right side of the image. This is certainly not the primary concept $\phi_{22}$ detects as it mainly detects blotted textures, but it certainly carries information about head location to the decoder. 

Fig. \ref{ex_gbp} depicts example class-attribute pairs where input attribution contributes in understanding of attributes. We use Guided Backpropagation \cite{guidedbackprop} (GBP) as input attribution method for ResNet on QuickDraw. It mainly assists in adding more support to our previously developed understanding of attributes. For eg., analyzing $\phi_5$ (relevant for Dog, Lion) based on AM+PI outputs suggested that it mainly detects curves similar to dog ears. The GBP output support this understanding as the most salient regions of the map correspond to curves similar to dog ears.

\subsection{Baseline implementations}
\label{baselines}

We cover the implementation details of various baselines used in this work (Tab 2, 3, 4 from main paper). As stated in the main paper, implementation of our method is available on Github \footnote{\url{https://github.com/jayneelparekh/FLINT}}.The accuracy of FLINT-$f$ is compared against BASE-$f$, PrototypeDNN, SENN. Fidelity of FLINT-$g$ is compared against VIBI and LIME.

\paragraph{BASE-$f$} We compare accuracy of FLINT-$f$ with BASE-$f$. The BASE-$f$ model has the same architecture as FLINT-$f$ but is trained with $\beta, \gamma, \delta = 0$, that is, only with the loss $\bmL_{pred}$ and not interpretability loss term. All the experimental settings while training this model are same as FLINT. 

\paragraph{PrototypeDNN} We directly report the accuracy of PrototypeDNN on MNIST, FashionMNIST (Tab 2 main paper) from the results mentioned in their paper \cite{protodnn}. Note that we do not report any results of PrototypeDNN on CIFAR10 and QuickDraw. This is because for processing more complex images and achieving higher accuracy, one would need to non-trivially modify architecture of their proposed model. Thus to avoid any unfair comparison, we did not report this result. The results of BASE-$f$ and SENN on CIFAR, QuickDraw help validate performance of FLINT-$f$ on QuickDraw.

\paragraph{SENN} We compare the accuracy as well as conciseness curve for FLINT with Self-Explaining Neural Networks (SENN) \cite{senn}. We implemented it with the help of their official implementation available on GitHub \footnote{\url{https://github.com/dmelis/SENN}}. SENN employs a LeNet styled network for MNIST in their paper. We use the same architecture for MNIST and FashionMNIST. For QuickDraw and CIFAR10 we use the VGG based architecture proposed for SENN in their paper to process more complex images. However, to maintain fairness, the number of attributes used in all the experiments for SENN are same as those for FLINT, that is, 25 for MNIST \& FashionMNIST, 24 for QuickDraw and 36 for CIFAR10, and also train for the same number of epochs. We use the default choices in their implementation for all hyperparameters and other settings. Another notable point is that although interpretations of SENN are worse than FLINT in conciseness (even when compared non-entropy version of FLINT), the strength of $\ell_1$ regularization in SENN is 2.56 times our strength (for identical $\bmL_{pred}$, i.e, cross-entropy loss with weight 1.0).

\paragraph{VIBI \& LIME} We benchmark the fidelity of interpretations of FLINT-$g$ for both by-design and post-hoc interpretation applications against a state-of-the-art black box explainer variational information bottleneck for interpretation (VIBI) \cite{vibi} and traditional explainer LIME \cite{lime}. Note that VIBI also possesses a model approximating the predictor for all samples. Both methods are implemented using the official repository for VIBI \footnote{\url{https://github.com/SeojinBang/VIBI}}. We compute the "\textit{Approximator Fidelity}" metric as described in their paper, for both systems. In the case of VIBI, this metric exactly coincides with our definition of fidelity. We set the hyperparameters to the setting that yielded best fidelity for datasets reported in their paper. For VIBI, chunk size $4 \times 4$, number of chunks $k=$ 20, for LIME, chunk size $2 \times 2$, number of chunks $k=$ 40. The other hyperparameters were the default parameters in their code.

\subsection{Subjective evaluation details}
\label{sub_eval}
The form taken by the participants can be accessed here \footnote{\url{https://forms.gle/PW6DEPZSmXb46Lnv9}}. 17 of the 20 respondents were in the age range 24-31 and at least 16 had completed a minimum of masters level of education in fields strongly related to computer science, electrical engineering or statistics. The form consists of a description where the participants are briefly explained through an example the various information (class-attribute pair visualizations and textual description) they are shown and the response they are supposed to report for each attribute, which is the level of agreement/disagreement with the statement: ``The patterns depicted in AM + PI outputs can be meaningfully associated to the textual description". As mentioned in the main paper, four descriptions (questions \#2, \#5, \#8, \#9 in the form) were manually corrupted to better ensure that participants are informed about their responses. The corruption mainly consisted of referring to other parts or concepts regarding the relevant class which are \textit{not} emphasized in the AM+PI outputs.

\section{Post-hoc interpretations}

\subsection{Implementation details}
The network architecture, the optimization procedures and hyperparameters are set to exactly the same values they were for their 'by-design', with one small change, $\beta$ for CIFAR10 is used as 0.3, and not 0.6, this is because for $\beta=0.6$, the system was running into scenario discussed in Sec.~\ref{hyperparams_text}, thus $\beta$ was lowered. 

\begin{figure}[!htb]
    \centering
    \includegraphics[width=0.48\textwidth]{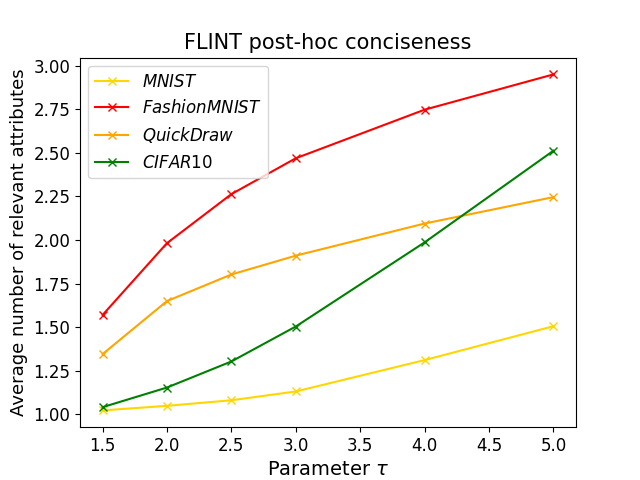}
    \caption{Conciseness curve of post-hoc interpretations generated using FLINT}
    \label{cns_ph}
\end{figure}

\paragraph{Results.} Fidelity benchmarked against VIBI is tabulated in Tab. \ref{fidelity_post-hoc} and conciseness curves for post-hoc interpretations are shown in Fig. \ref{cns_ph}. They clearly indicate that FLINT can yield high fidelity and highly concise \textit{post-hoc} interpretations.
\begin{table}
\footnotesize
\centering
\begin{tabular}{l c c} 
\toprule
Dataset & VIBI & FLINT-$g$\\
\toprule
MNIST & 95.8$\pm$0.2 & \textbf{98.6$\pm$0.2}\\
FashionMNIST & 88.4$\pm$0.2 & \textbf{92.8$\pm$0.3} \\ 
CIFAR10 & 64.2$\pm$0.3 & \textbf{89.1$\pm$0.5}\\
QuickDraw & 78.0$\pm$0.4 & \textbf{90.5$\pm$0.3}\\
\bottomrule
\end{tabular}
\caption[Results]{
Fidelity for post-hoc interpretations of BASE-$f$ (in \%)
}
\label{fidelity_post-hoc}
\end{table}

\subsection{Additional visualizations}

\begin{figure}[!htb]
\centering
\includegraphics[width=0.6\textwidth]{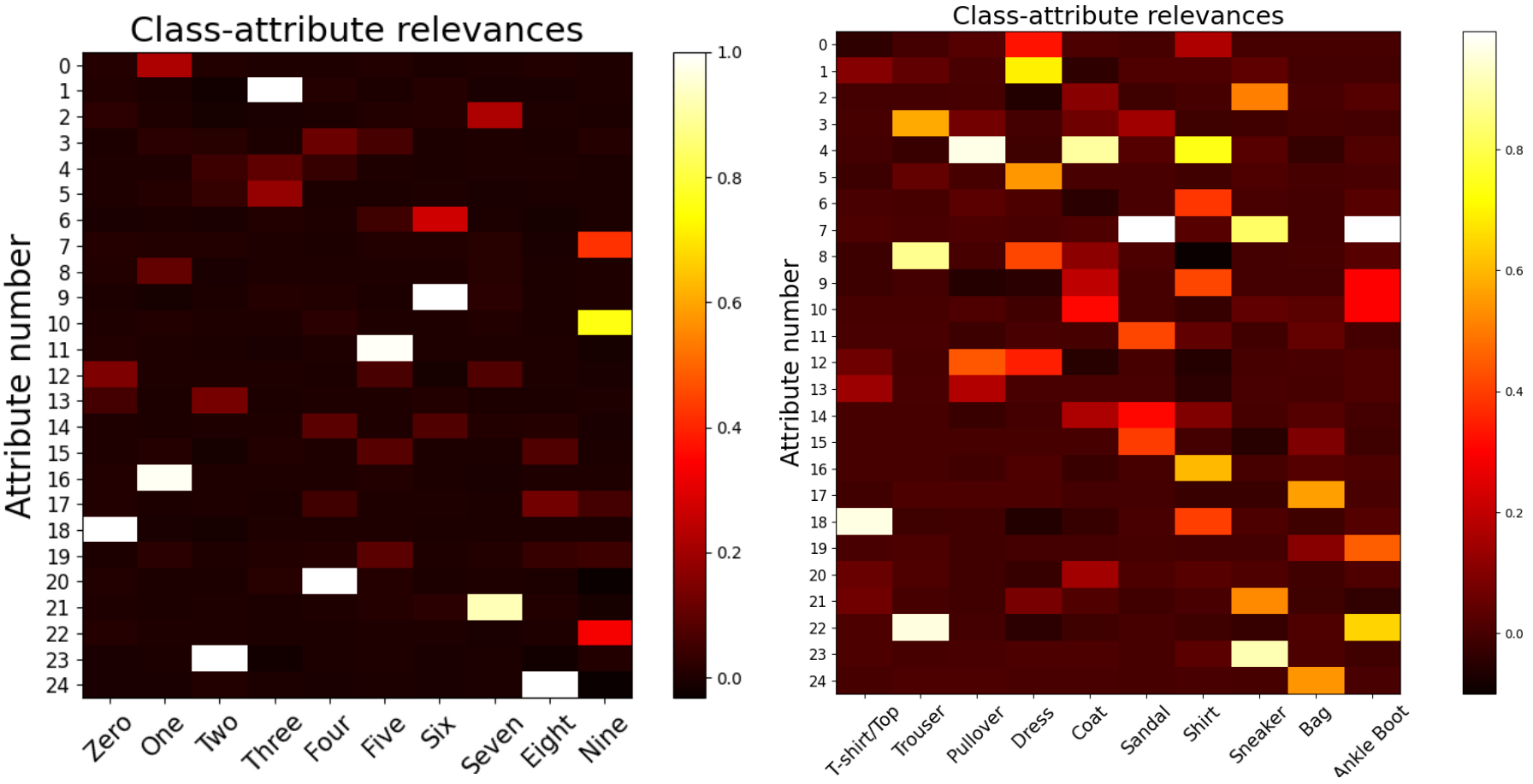}
\caption{Global class-attribute relevances $r_{j,c}$ for post-hoc interpretations on MNIST (Left) and FashionMNIST (Right). 15 class-attribute pairs for MNIST and 28 pairs for FashionMNIST have relevance $r_{j,c} > 0.2$.}
\label{global_rel1_sup_ph}
\end{figure}

\begin{figure}[!htb]
\centering
\includegraphics[width=0.6\textwidth]{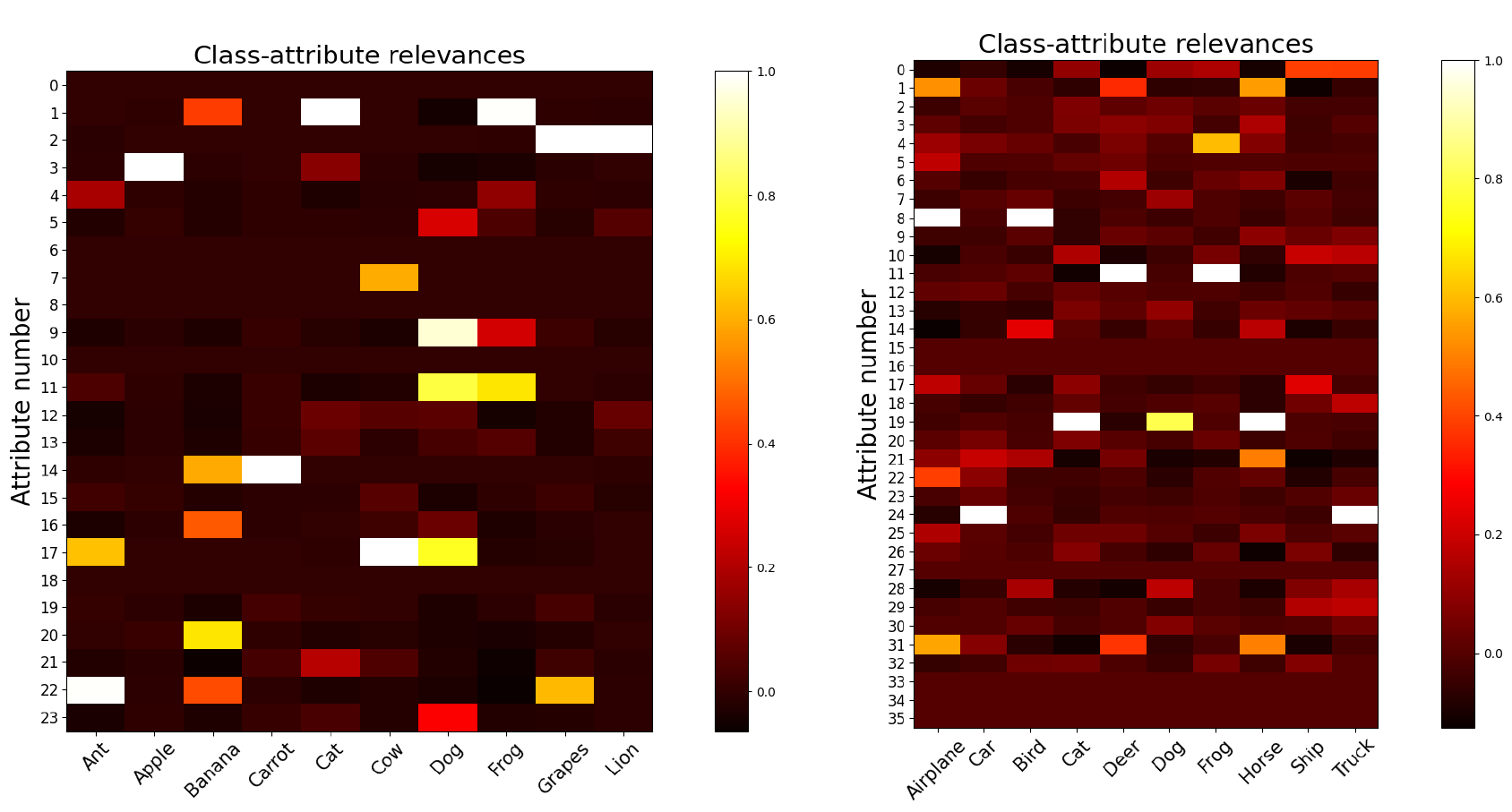}
\caption{Global class-attribute relevances $r_{j,c}$ for post-hoc interpretations on QuickDraw (Left) and CIFAR10 (Right). 24 class-attribute pairs for QuickDraw and 26 pairs for CIFAR10 have relevance $r_{j,c} > 0.2$.}
\label{global_rel2_sup_ph}
\end{figure}

\begin{figure}[!htb]
\centering
\includegraphics[width=0.6\textwidth]{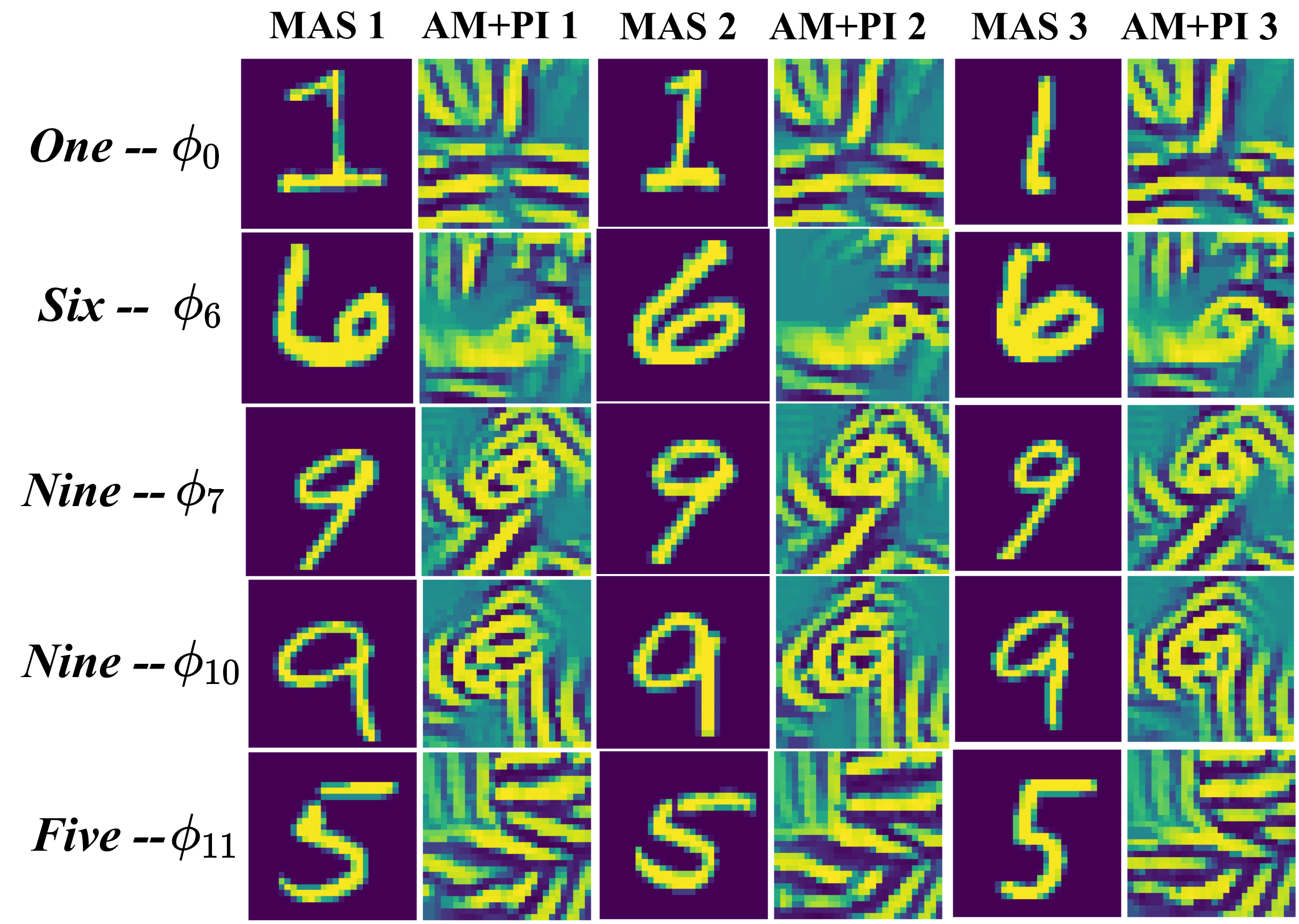} 
\caption{Sample class-attribute visualizations for post-hoc interpretations for MNIST.}
\label{mnist_global_viz_ph}
\end{figure}

\begin{figure}[!htb]
\centering
\includegraphics[width=0.6\textwidth]{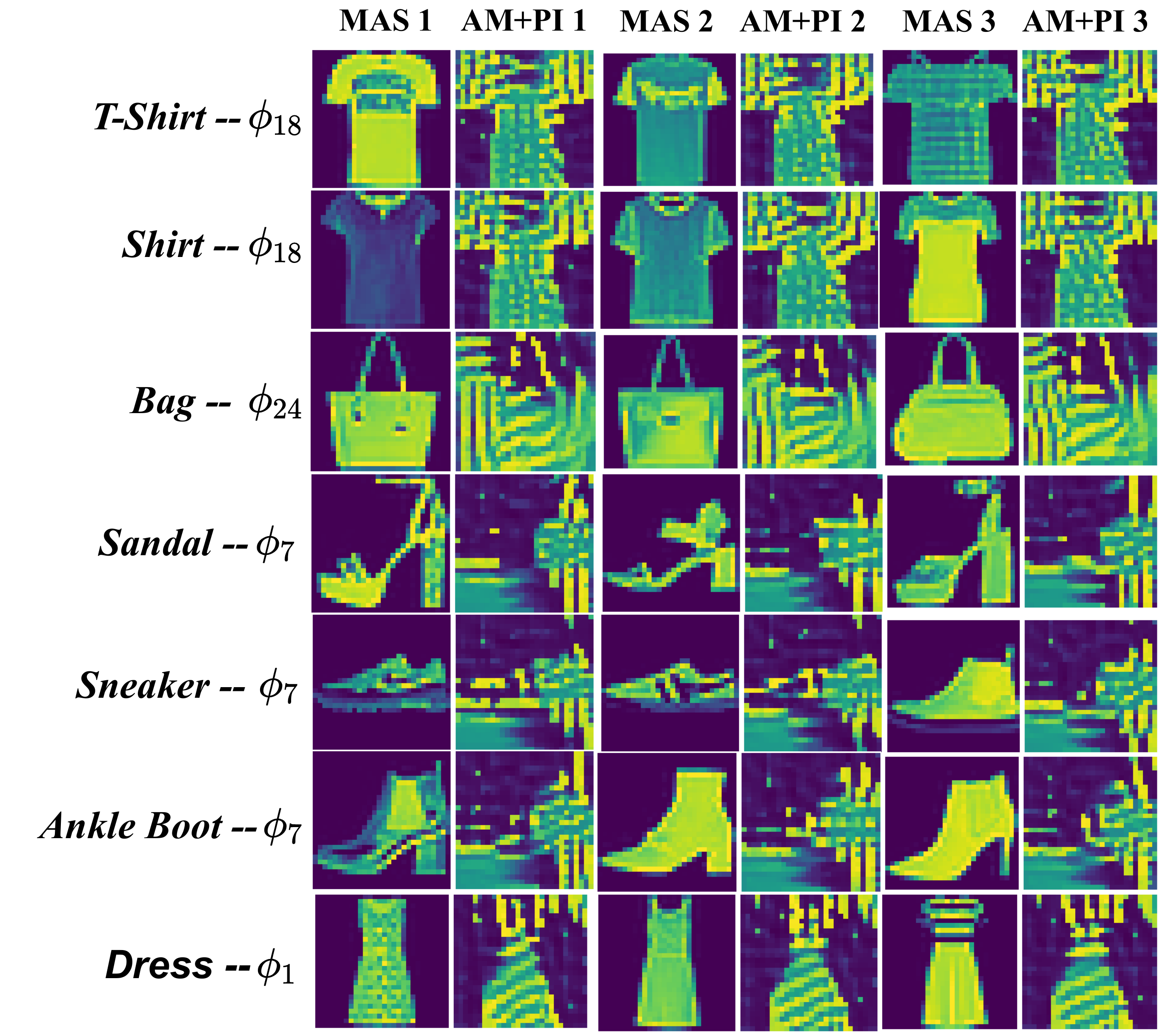} 
\caption{Sample class-attribute visualizations for post-hoc interpretations for Fashion-MNIST}
\label{fmnist_global_viz_ph}
\end{figure}

\begin{figure}[!htb]
\centering
\includegraphics[width=0.6\textwidth]{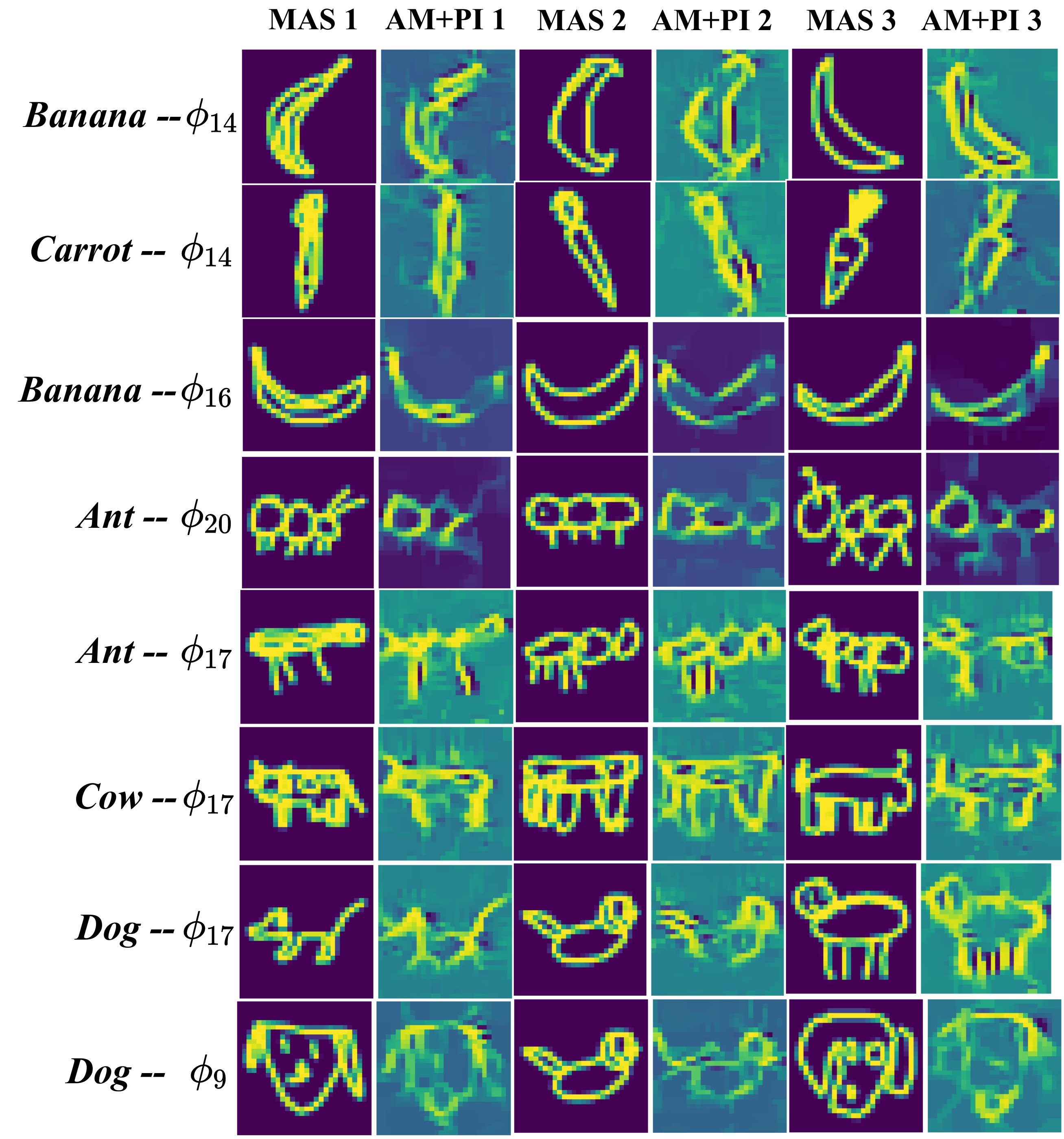} 
\caption{Sample class-attribute visualizations for post-hoc interpretations on QuickDraw}
\label{qdraw_global_viz_ph}
\end{figure}

\begin{figure}[!htb]
\centering
\includegraphics[width=0.56\textwidth]{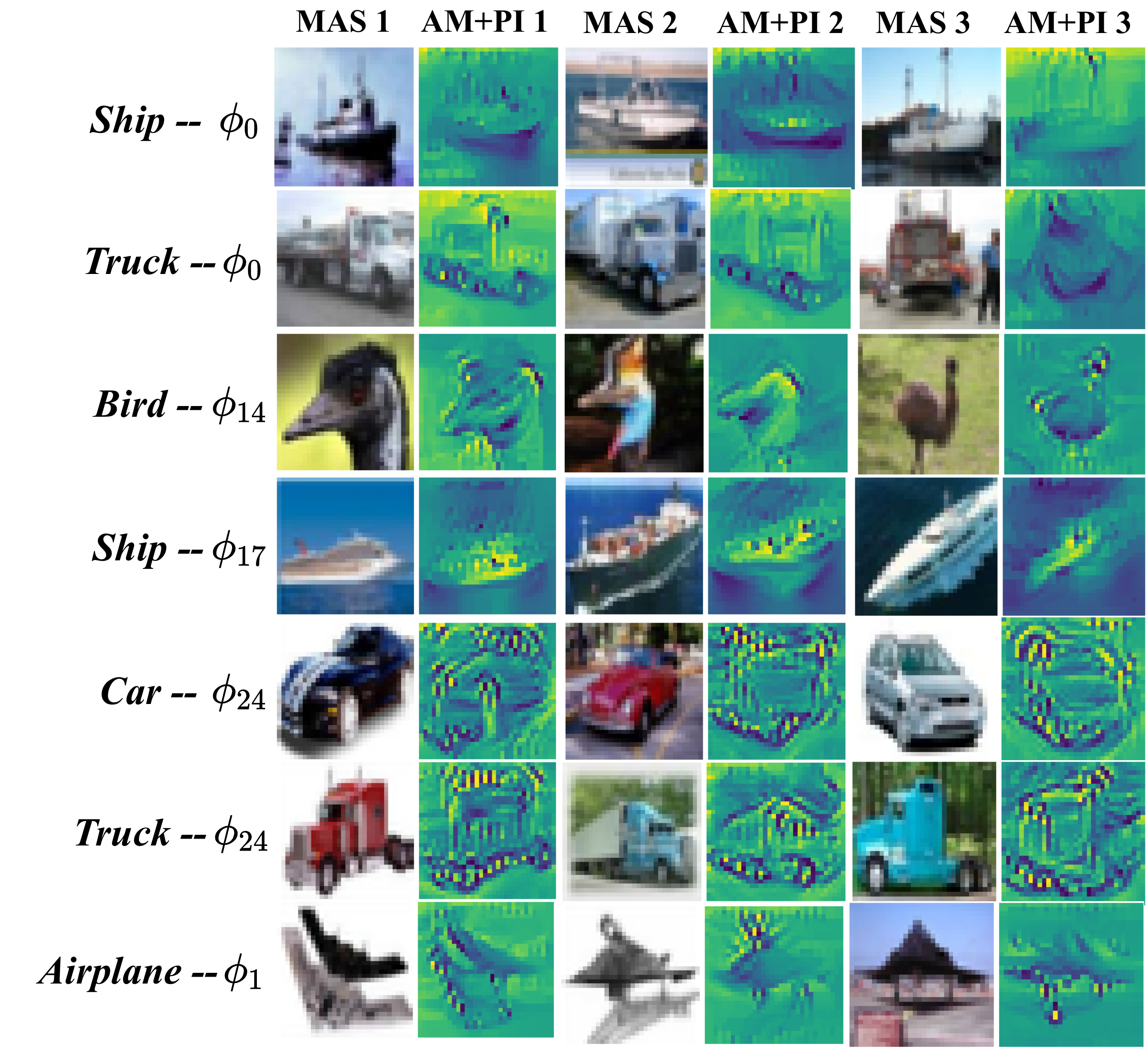} 
\caption{Sample class-attribute visualizations for post-hoc interpretations on CIFAR-10}
\label{cifar10_global_viz_ph}
\end{figure}

Figs. \ref{global_rel1_sup_ph} and \ref{global_rel2_sup_ph} contain global relevances for post-hoc interpretations on all four datasets. Figs. \ref{mnist_global_viz_ph}, \ref{fmnist_global_viz_ph}, \ref{qdraw_global_viz_ph} and \ref{cifar10_global_viz_ph}, illustrate some additional visualizations of class-attribute pairs on all datasets.

\subsection{Experiments using ACE}
\label{ace_exp}
We conducted additional experiments using ACE to interpret trained models from our experiments. The key bottleneck for ACE's application on our datasets and networks is the use of CNN as a similarity metric (to automate human annotation) for image segments irrespective of their scale, aspect ratio. This is a specialized property only been empirically shown for specific CNN's trained on ImageNet (as discussed in their paper). The networks trained on our datasets thus very often cluster unrelated segments, resulting in little to no consistency in any extracted concept. To illustrate the above we describe the experimental settings and show extracted concepts for a few classes from QuickDraw and CIFAR-10 on the BASE-$f$ models. The quality of results is the same when interpreting FLINT-$f$ models although we only illustrate interpretations from BASE-$f$ models.

\paragraph{Experimental setting.} We utilize the official open-sourced implementation of their method \footnote{\url{https://github.com/amiratag/ACE}}. Due to the smaller sized images we perform segmentation at a single scale. We experimented with different configurations for ``number of segments" and ``number of clusters/concepts". The number of segments were varied from 3 to 15. For higher values the segments were often too small for concepts to be meaningful. We thus kept the number of segments 5 for each sample. For each class we chose 100 samples. The number of clusters were varied from 5 to 25. Due to the smaller number of segments (compared to original experiments from ACE which used 25), we kept number of clusters at 12. We access the deepest intermediate layer used in experiments with FLINT (shown in Fig. \ref{arch_2}). 

\begin{figure}[t!]
\centering
\includegraphics[width=1.05\textwidth]{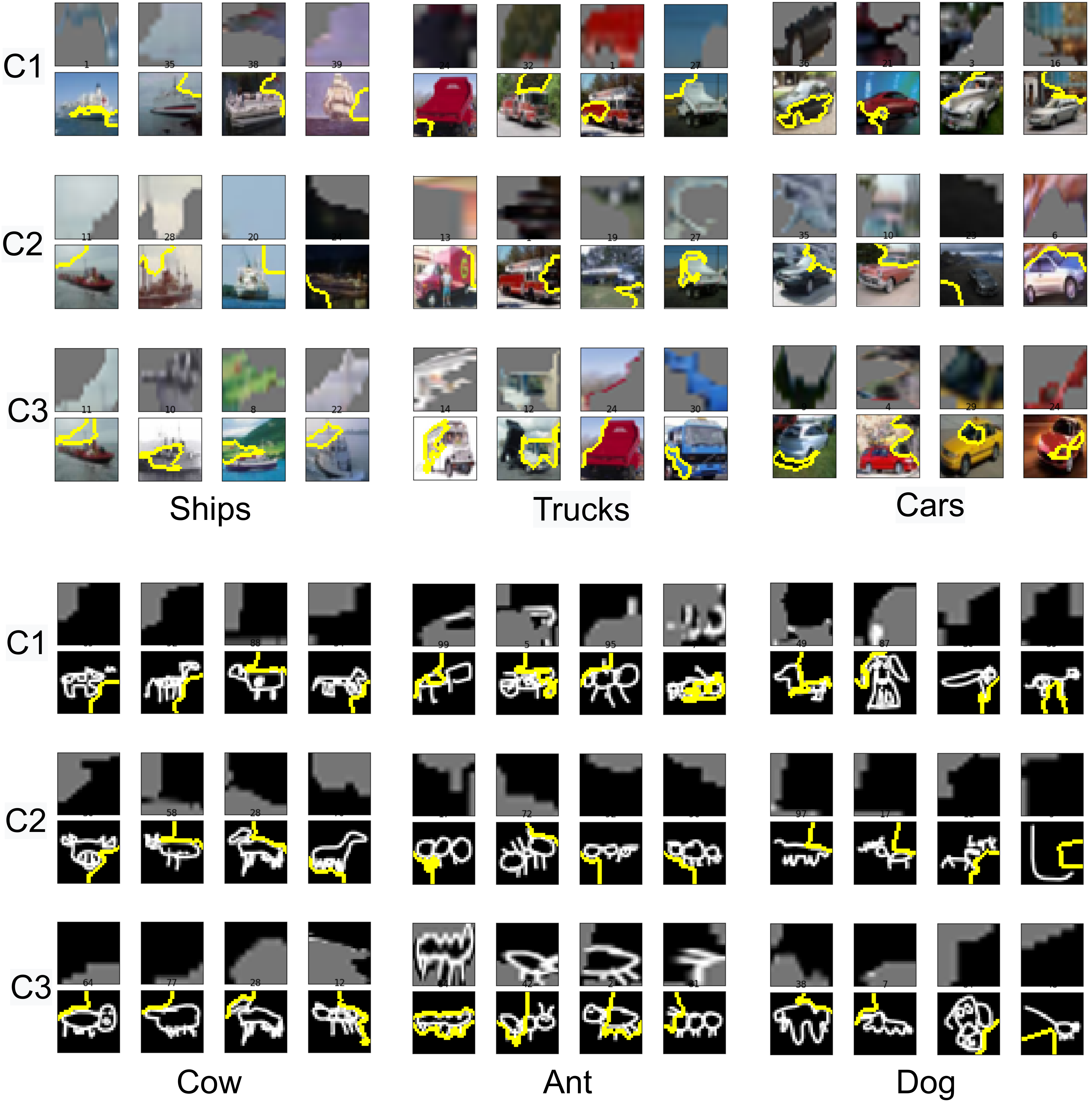} 
\caption{Discovered concepts using ACE for 3 classes on CIFAR-10 (Top) and QuickDraw (Bottom). We show the top 3 concepts according to their TCAV scores. Each concept consists of 4 segments extracted from images of the class. They are shown in 2 rows, the first contains the segments and the second shows where the segment was extracted from.}
\label{ace_results}
\end{figure}

\paragraph{Results.} The top 3 discovered concepts (according to the TCAV scores) are shown in Fig. \ref{ace_results}. The segments for any concept on CIFAR show almost no consistency. This is mainly because the second step pf ACE, requiring a CNN's intermediate representations to replace a human subject for measuring the similarity of superpixels/segments, is hard to expect for these networks not trained on ImageNet. Thus, segments capturing background or any random part of the object, completely unrelated, end up clustered together. For QuickDraw, the segmentation algorithm also suffers problems in extracting meaningful segments due to sparse grayscale images. It generally extracts empty spaces or a big chunk of the object itself. This, compounded with the earlier issue about segment similarity results in mostly meaningless concepts. The only slight exception to this is concept 3 for 'Ant' for which two segments capture a single flat blob with small tentacles.       

\section{Limitations}

\begin{itemize}
    \item The current design of attributes and their encoded concept visualization procedure is more suited for classification tasks and image as input modality. Although multiple proposed losses/visualization tools could be generalized to other input modalities (e.g. audio, video, graphs etc.) or other machine learning tasks (regression), it requires work in that direction. 
    
    \item The set of proposed properties is not exhaustive and can be further improved. It could be desired that attributes encode concepts which are invariant to certain transformations, or focus on specific spatial regions, or are robust to adversarial attacks / specific types of noise or contamination. 
    
    

    \item The choice of hidden layers requires some level of experience with neural architectures.
\end{itemize}

\section{Potential negative societal impact}

Interpretability becoming a frequently raised issue when training and exploiting neural network (NN) architectures, the main expected societal impact of FLINT is improvement of their understandability as well as providing explanations of the decisions made by NNs. Nevertheless, even this intrinsically benevolent machinery can be used for harm when in malicious hands.

Potential misuse can be expected on two different levels: First, if incorrectly trained (e.g., wrong NN design, insufficient number of training examples and/or or training epochs, in particular for FLINT-$f$), due to lack of knowledge or on purpose, FLINT can provide misleading interpretations. Second, even a well-trained explainable AI can serve evil purpose in hands of a maliciously destined user.

Clearly, the authors expect proper use of the developed FLINT methodology, although direct misuse-protection mechanisms were not developed in this piece of research, not being the initial goal.


\end{document}